\documentclass{naturep}

\usepackage{amssymb}
\usepackage{amsmath}
\usepackage{graphicx}
\usepackage{algorithm}
\usepackage{algpseudocode}
\usepackage{ltablex}

\usepackage{bbm}
\usepackage{lineno}
\usepackage[margin=0.6in]{geometry}
\usepackage{ragged2e}
\usepackage{setspace}
\usepackage{longtable}
\usepackage{hyperref}
\usepackage{anyfontsize}
\usepackage{setspace}
\usepackage{float}
\usepackage{booktabs}
\usepackage{multirow}
\usepackage{xcolor}
\usepackage{blindtext}

\usepackage{caption}
\usepackage{chngcntr}
\usepackage{subcaption}

\usepackage{cleveref}

\newcommand{\mainref}[1]{\textbf{\Cref{#1}}}
\newcommand{\edref}[1]{\textbf{Extended Data~\Cref{#1}}}
\makeatletter
\let\saved@includegraphics\includegraphics
\AtBeginDocument{\let\includegraphics\saved@includegraphics}

\makeatother

\usepackage{enumitem, kantlipsum}

\title{\begin{flushleft}{\begin{spacing}{1} A Clinical-Grade Agentic and Generative AI-driven Copilot for Human Pathology\end{spacing}}\end{flushleft}}

\title{\begin{flushleft}{\begin{spacing}{1} Evidence-based diagnostic reasoning with multi-agent copilot for human pathology\end{spacing}}\end{flushleft}}

\newcommand{\agent}{SlideSeek}
\newcommand{\pathchatnew}{PathChat+}
\usepackage{xcolor}
\newcommand{\kuanedit}[1]{\textcolor{black}{#1}}
\newcommand{\lucaedit}[1]{\textcolor{black}{#1}}

\begin{document}

\maketitle
\begin{spacing}{1}
    
\vspace{-16mm}
\noindent Luca L. Weishaupt$^{1,2,\boldsymbol{\ddag}}$, Chengkuan Chen$^{1,\boldsymbol{\ddag}}$, Drew F. K. Williamson$^{1,3,*}$, Richard J. Chen$^{1,2,3,*}$, Guillaume Jaume$^{1,3,4,*}$, Tong Ding$^{1,5,*}$, Bowen Chen$^{1,2,*}$, Anurag Vaidya$^{1,2}$, Long Phi Le$^{2}$, Ming Y. Lu$^{1,2,3,6+}$, 
and Faisal Mahmood$^{1,3,4,7,+}$
\end{spacing}
\vspace{-9mm}
\begin{spacing}{1.4}
\begin{affiliations}
 \item Department of Pathology, Brigham and Women's Hospital, Harvard Medical School, Boston, MA
 \item Health Sciences and Technology, Harvard-MIT, Cambridge, MA
 \item Department of Pathology, Massachusetts General Hospital, Harvard Medical School, Boston, MA
 \item Cancer Program, Broad Institute of Harvard and MIT, Cambridge, MA 
 \item Harvard John A. Paulson School of Engineering and Applied Sciences, Harvard University, Cambridge, MA
 \item Electrical Engineering and Computer Science, Massachusetts Institute of Technology (MIT), Cambridge, MA

 \item Harvard Data Science Initiative, Harvard University, Cambridge, MA
 \\$\boldsymbol{\ddag}$ Co-1st authors, $*$ Co-2nd authors, $+$ Co-senior authors
 \\\textbf{Lead contact}: Faisal Mahmood (faisalmahmood@bwh.harvard.edu)
 
\end{affiliations}
\end{spacing}

\vspace{-8mm}
\begin{spacing}{1}

\noindent \textbf{Abstract}\\
\lucaedit{
Pathology is undergoing rapid transformation through whole-slide imaging and artificial intelligence (AI)\cite{madabhushi2016image, bera2019artificial, heinz2022future, cui2021artificial, abels2019computational}. While advances in multimodal large language models (LLMs) have shown promise in classifying histology images and answering questions, they remain constrained to small regions of interest (ROIs) and lack the ability to autonomously reason across gigapixel tissue slides or draft pathology-style reports\cite{pathchat, llava-med, quilt_llava, pa-llava}. Here, we present SlideSeek, a multi-agent copilot that autonomously explores whole-slide images through iterative, hierarchical reasoning. A supervisor agent generates diagnostic hypotheses and adaptively assigns tasks to explorer agents, which examine regions and return findings that the supervisor synthesizes into a visually-grounded report. At the foundation of SlideSeek lies PathChat+, a pathology-specific multimodal LLM trained on 1.1M instructions and 5.5M Q\&A turns, which achieves state-of-the-art captioning capabilities across ten ROI-level benchmarks. On DDxBench, a differential diagnosis benchmark spanning 55 diseases, SlideSeek achieved 86.0\% top-1 accuracy and 92.7\% top-3 accuracy, outperforming general-purpose multimodal LLMs by up to 42\% and maintaining strong performance on 41 rare diseases. Ablation studies confirmed the need for multi-agent reasoning and a strong morphological captioner such as PathChat+. Analysis of SlideSeek’s failure modes revealed challenges in fine-grained grading and the detection of small lesions. By combining autonomous navigation, interpretable reasoning, and domain-specific captioning with PathChat+, SlideSeek establishes a generalizable framework for agentic copilots in diagnostic pathology.
}
\end{spacing}

\vspace{-4mm}
\begin{spacing}{1}
\newpage
\noindent\textbf{\large{Introduction}} 

\noindent \lucaedit{Pathology is undergoing rapid digital transformation, driven by the adoption of whole-slide imaging and advances in artificial intelligence (AI)\cite{madabhushi2016image,bera2019artificial,heinz2022future,cui2021artificial,abels2019computational}. AI methods in pathology have achieved success in cancer subtyping and grading\cite{coudray2018classification,lu2021data,bulten2020automated,nagpal2019development,huang2022deep,campanella2019clinical,bejnordi2017diagnostic}, biomarker discovery\cite{kather2020pan,fu2020pan,saldanha2023self,wagner2023transformer}, outcome prediction\cite{beck2011systematic,mobadersany2018predicting,porpoise,lee2022derivation,courtiol2019deep,lu2020prognostic,amgad2023population,boehm2022multimodal,sammut2022multi,vanguri2022multimodal,huang2023artificial}, among other tasks\cite{lu2021ai,zhu2023accurate,sish,yottixel,smily,wang2023retccl,yala2022optimizing,zhou2023multi,laleh2022benchmarking,graham2019hover,graham2023one}. Yet these systems are typically developed under narrow, task-specific supervision, where models are optimized to recognize predefined categories, such as breast cancer subtypes\cite{pati2022hierarchical} or Gleason grades\cite{bulten2020automated}. Such close-context training constrains their ability to generalize across organs, disease spectra, and laboratory settings, and prevents them from incorporating broader clinical context or reasoning across the gigapixel scale of whole-slide images. This stands in contrast to human pathologists, who integrate slide-wide morphology with clinical context when making diagnostic decisions\cite{bulten2020automated,huang2022deep,campanella2019clinical,lu2021data}. }

\noindent \lucaedit{Foundation models\cite{vorontsovFoundationModelClinicalgrade2024,wangPathologyFoundationModel2024,xiangVisionLanguageFoundation2025,xuWholeslideFoundationModel2024,tuGeneralistBiomedicalAI2024} and multimodal large language models (LLMs) have begun to address this gap by extending beyond purely supervised visual analysis and enabling ``zero-shot'' classification from written text prompts\cite{huang2023artificial,lu2023visual,quilt,biomedclip,pathasst,quilt_llava,pa-llava,pathchat,llava-med,sunPathGen16M16Million2024,singhalLargeLanguageModels2023}. However, their utility in pathology remains narrowly explored. Most evaluations focus on isolated regions of interest (ROIs), leaving their ability to integrate findings across regions or to provide slide-level diagnostic reasoning untested. Moreover, these systems remain reactive and are designed to address queries without explicit mechanisms for proactive planning or exploration\cite{o1,deepseek,talebiradMultiAgentCollaborationHarnessing2023,swansonVirtualLabAI2025,schmidgallAgentClinicMultimodalAgent2025,luAIScientistFully2024,ghezlooPathFinderMultiModalMultiAgent2025,gaoEmpoweringBiomedicalDiscovery2024,ferberDevelopmentValidationAutonomous2025}. This limits their clinical applicability, where diagnoses often hinge on detecting subtle, spatially localized features and synthesizing evidence across large tissue areas. Early attempts at learned navigation and zooming\cite{zoommil,pathfinder,buzzard2024paths,wang2025pathology} represent progress, but they lack explicit architectures for multi-step reasoning and often continue to rely on task-specific supervision, limiting their generalization across organs and diseases.}  

\begin{figure*}[th!]
\centering
\includegraphics[width=\textwidth]{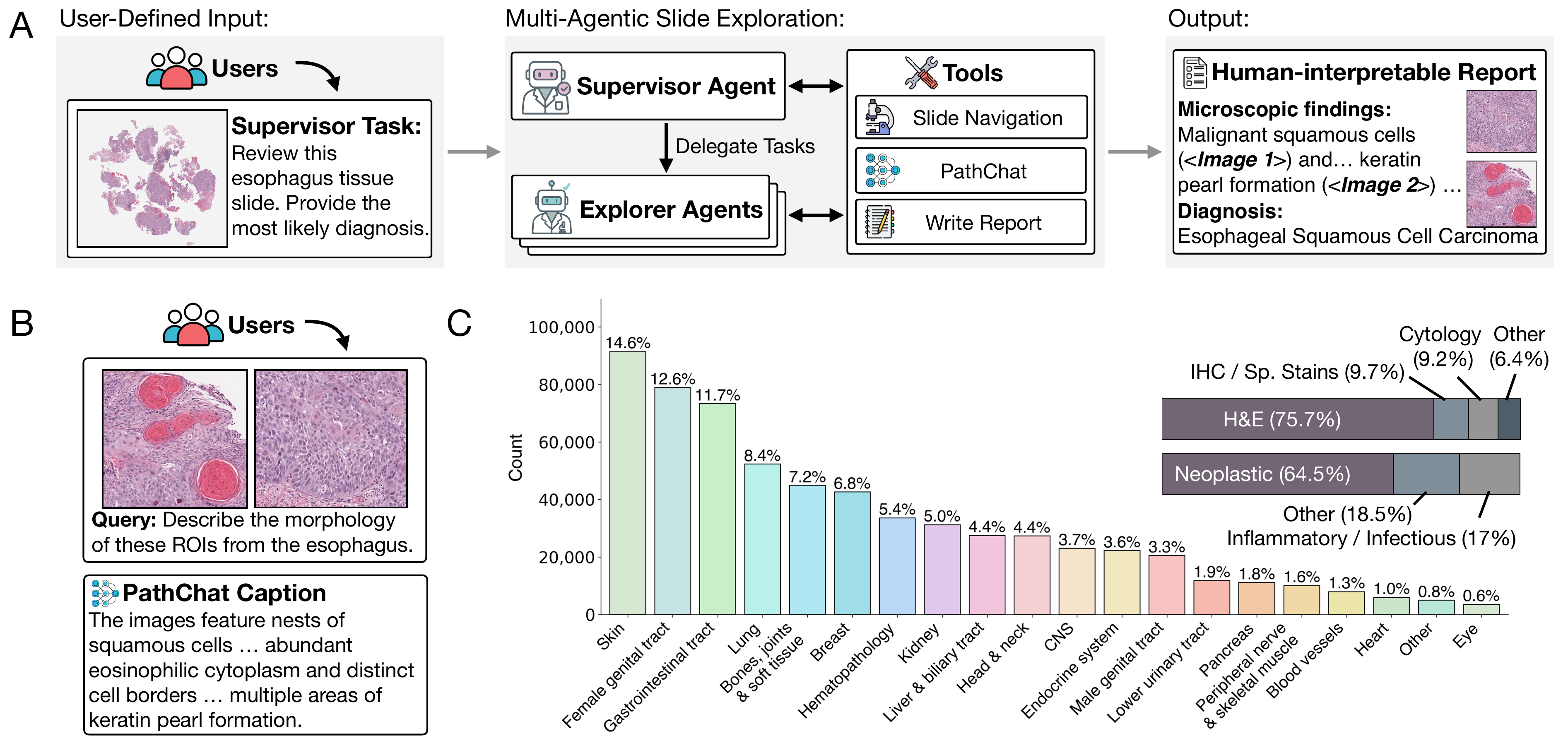}
\vspace{-.5cm}
\caption{\textbf{\lucaedit{Overview of \agent{} and \pathchatnew{} for slide-level diagnosis and report generation}.}
\lucaedit{\textbf{A.} Our multi-agent-based AI system, \agent{}, starts with a standardized task description to reach a slide diagnosis autonomously. A reasoning large language model (LLM) serves as supervisor, continually tracking progress, refining diagnostic plans, and choosing additional regions for morphological examination. During each planning iteration, the supervisor instructs a team of specialized pathologist agents, each of which interacts with \pathchatnew{} to analyze specific regions and report findings back to the supervisor. This iterative, hierarchical workflow continues until the supervisor agent determines sufficient evidence has been collected to establish a well-supported differential diagnosis. A separate report agent synthesizes the morphological evidence from critical ROIs into an interpretable, visually grounded diagnostic summary report. See \edref{fig:SlideSeekArch} for a step-by-step view of the supervisor–explorer workflow.
\textbf{B.} Example of \pathchatnew{} that can intake one or multiple regions of interest (ROIs) with text instructions to provide morphologically-grounded tissue description. 
\lucaedit{\textbf{C.}} \pathchatnew{} is trained with instruction finetuning on 1.13 million instructions and 5.49 million question-answer turns, based on 624 thousand unique images. A distribution of images is provided by tissue site, staining, and disease category.}}
\label{fig:fig1}
\end{figure*}

\noindent \lucaedit{To address these gaps, we introduce \agent{}, a reasoning-enabled multi-agent system for autonomous assessment of gigapixel whole-slide images. As illustrated in \textbf{Figure \ref{fig:SlideSeekArch}A}, given a diagnostic task, \agent{} executes a supervisor–explorer loop that plans, hierarchically queries, and iteratively inspects candidate regions, integrating multi-scale evidence into a slide-level diagnosis and structured pathology report. At its core, \agent{} builds on \pathchatnew{}, a powerful pathology image analysis copilot tool. \pathchatnew{} is a pathology-specific multimodal LLM, trained on over one million visual–language instruction samples, five million question-and-answer turns spanning 624 thousand images across diverse organs, tissues, and diseases (\textbf{Figure \ref{fig:fig1}B-C}). \pathchatnew{} builds upon its predecessor, PathChat 1\cite{pathchat}, with additional support for high-resolution multi-ROI analysis, enabling it to provide detailed morphological interpretation of larger tissue regions.  }

\noindent \lucaedit{To evaluate our approach, we designed complementary experiments at two levels of analysis. First, we assess \agent{} on DDxBench, a benchmark of 150 whole-slide images covering 55 neoplastic diseases, including 41 classified as rare, to test its ability to generate open-ended differential diagnoses. Second, we benchmark \pathchatnew{} independently across ten ROI-level benchmarks spanning visual question answering, classification, and captioning tasks. This large-scale evaluation enables direct comparison against state-of-the-art multimodal LLMs, including closed-weight frontier models such as GPT-5, open-weight models such as Qwen3-VL, and specialized medical models such as Quilt-LLaVA\cite{quilt_llava}. These evaluations allow us to establish both the feasibility of autonomous slide-level reasoning with \agent{} and the importance of pathology-specific multimodal pretraining for high-fidelity morphological interpretation with \pathchatnew{}.}

\captionsetup[figure]{
  name=Figure,
  labelfont=bf, 
}
\clearpage

\noindent\textbf{\large{Results}}

\noindent\textbf{ROI-level benchmarking across diverse tasks}

\begin{figure*}[h!]
\centering
\includegraphics[width=\textwidth]{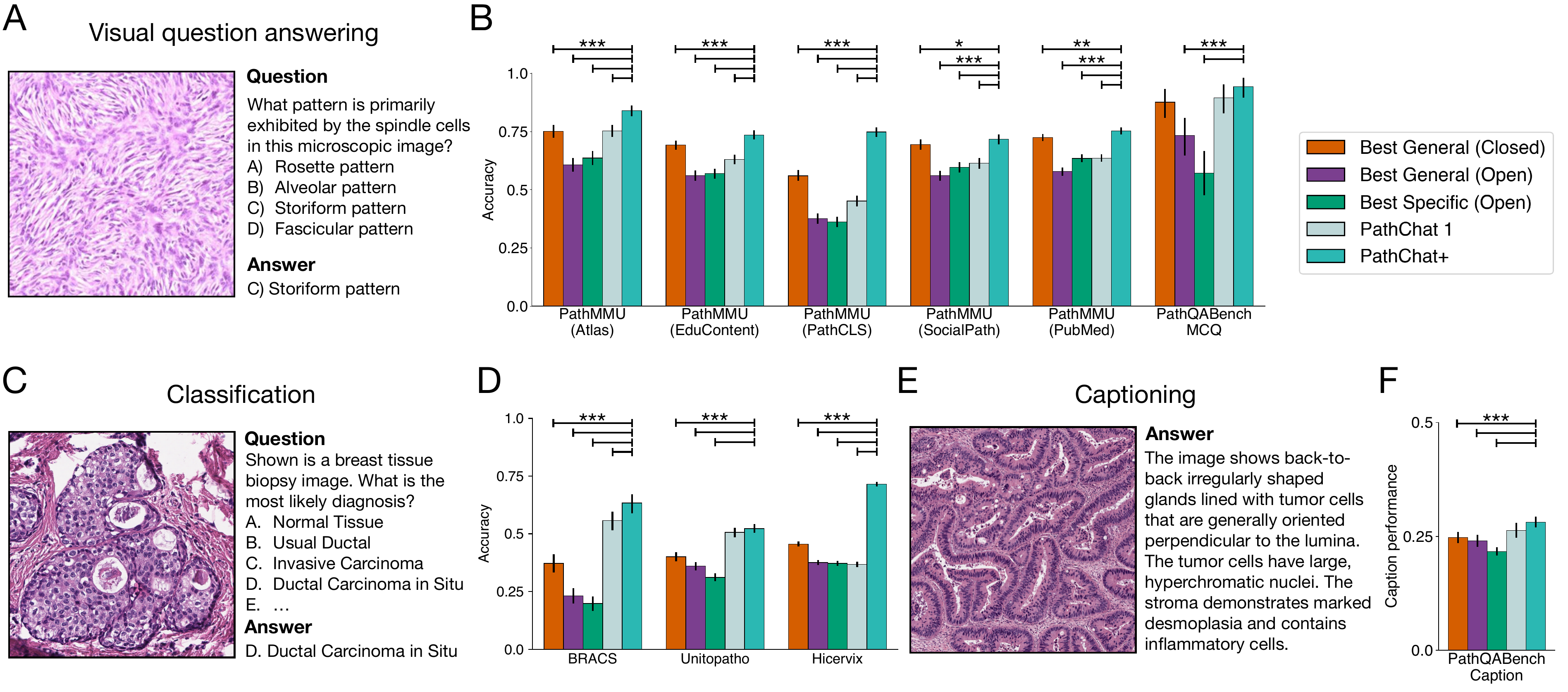}
\caption{\lucaedit{\textbf{ROI-level benchmarking of \pathchatnew{} versus multimodal LLM baselines.}
\textbf{A.-B.} Visual question answering (VQA): \textbf{A.} example ROI with prompt and answer; \textbf{B.} accuracy on PathMMU subsets (Atlas, EduContent, PathCLS, SocialPath, PubMed) and PathQABench MCQ.
\textbf{C.-D.} Image classification: \textbf{C.} example multiple-choice prompt and ground truth; \textbf{D.} accuracy on BRACS, UniToPatho, and HiCervix.
\textbf{E.-F.} Captioning: \textbf{E.} example model-generated morphological description; \textbf{F.} METEOR score on PathQABench-Caption.
Bars report, for each family, the strongest closed-source general model (“Best General (Closed)”), strongest open-source general model (“Best General (Open)”), and strongest medical-specialized model (“Best Specific (Open)”), alongside PathChat 1 and \pathchatnew{} (legend). Full per-model results are in \edref{tab:PathMMU_all,tab:PathMMU_Atlas,tab:PathMMU_EduContent,tab:PathMMU_PathCLS,tab:PathMMU_SocialPath,tab:PathMMU_PubMed,tab:becnchmark_PathQABench,tab:benchmark_classification,tab:becnchmark_PathQABenchCaption}. Error bars denote 95\% confidence intervals from non-parametric bootstrapping. In \textbf{B.}, \textbf{D.}, and \textbf{F.}, statistical significance was determined between \pathchatnew{} compared to all other models using a paired two-sided permutation test ($n=1000$). The p-values are indicated as $p<0.05$: *, $p<0.01$: **, $p<0.001$: ***.
Statistical }}
\label{fig:ROI-bench}
\end{figure*}

\noindent \lucaedit{We first evaluated the standalone capabilities of \pathchatnew{} across three core pathology tasks: visual question answering (VQA), image classification, and image captioning (\textbf{Figure \ref{fig:ROI-bench}A}). For VQA, we used the five expert-validated benchmarks from PathMMU\cite{sun2024pathmmu}, which span multiple organ types and pathology subspecialties, as well as PathBenchQA MCQ\cite{pathchat}, which focuses on diagnostic reasoning from ROI images paired with clinical context. For classification, we included three datasets, BRACS (breast neoplasm subtyping), UniToPatho (colorectal polyp assessment), and HiCervix (cervical cytology), reformatted into a multiple-choice setting. For captioning, we designed a captioning variant of the existing PathBenchQA\cite{pathchat} dataset, where board-certified pathologists annotated each image with a reference morphological description. Performance was compared against a set of 12 multimodal LLMs, including closed-weight frontier models (GPT-5\footnote{https://cdn.openai.com/pdf/8124a3ce-ab78-4f06-96eb-49ea29ffb52f/gpt5-system-card-aug7.pdf\label{gpt-5-fn}}, GPT-5-mini\footref{gpt-5-fn}, Claude Sonnet 4\footnote{https://www-cdn.anthropic.com/07b2a3f9902ee19fe39a36ca638e5ae987bc64dd.pdf}, Gemini 2.5 pro\footnote{https://storage.googleapis.com/model-cards/documents/gemini-2.5-pro.pdf}), open-weight general models (Qwen3-VL\cite{qwen3technicalreport}, Llama 3.2\cite{llama3} and LLaVA-OneVision\cite{llava-ov}), and specialized medical/pathology models (HuatuoGPT-Vision\cite{chen2024huatuogptvisioninjectingmedicalvisual}, LLaVA-Med 1.5\cite{llava-med}, PA-LLaVA\cite{pa-llava}, Quilt-LLaVA\cite{quilt_llava}, and the first iteration of PathChat)\cite{pathchat}. For clarity, \textbf{Figure \ref{fig:ROI-bench}B} reports the strongest model from each category. Per-subset results are summarized in \edref{tab:PathMMU_all} with detailed breakdowns in \edref{tab:PathMMU_Atlas,tab:PathMMU_EduContent,tab:PathMMU_PathCLS,tab:PathMMU_SocialPath,tab:PathMMU_PubMed}, PathQABench MCQ results in \edref{tab:becnchmark_PathQABench}, classification results in \edref{tab:benchmark_classification} and captioning results in \edref{tab:becnchmark_PathQABenchCaption}.}

\noindent In visual question answering, \pathchatnew{} achieved the highest performance across all six benchmarks (\textbf{Figure \ref{fig:ROI-bench}B}), surpassing both frontier general-purpose models and specialized medical multimodal LLMs, as well as its predecessor PathChat 1\cite{pathchat}. \lucaedit{On the combined PathMMU test set, \pathchatnew{} improved absolute accuracy by an average of 10.6\% over frontier models ($p < 0.001$), with gains ranging from +2.3\% to +18.7\% across individual sources. On PathQABench MCQ (\edref{tab:becnchmark_PathQABench}), which emphasizes diagnostic reasoning, \pathchatnew{} outperformed Gemini 2.5 Pro by 6.7\% ($p = 0.099$). When compared to the best domain-specific models (i.e., HuatuoGPT-Vision), the margins were even larger, with an absolute performance gain of 18.7\% on PathMMU and 37.2\% on PathQABench MCQ ($p < 0.001$). A similar trend held for image classification, where \pathchatnew{} consistently outperformed all other types of models (\edref{tab:benchmark_classification}). For example, it outperformed Gemini 2.5 Pro by 26.1\% and HuatuoGPT-Vision by 43.5\% on the BRACS benchmark ($p < 0.001$). In captioning, \pathchatnew{} also set a new state of the art, achieving the highest METEOR\cite{banerjee2005meteor} score on PathQABench Caption relative to expert-annotated ground truth. Overall, \pathchatnew{} statistically outperforms its predecessor, PathChat 1 in 7/10 tasks ($p < 0.05$), and is statistically better than all other models in 10/10 tasks ($p < 0.05$, except comparing to Gemini 2.5 pro on PathQABench MCQ where $p=0.099$). Examples of generated morphological descriptions are included in \edref{fig:caption}.}

\noindent \lucaedit{Taken together, these results establish \pathchatnew{} as the strongest model across ten ROI-level benchmarks, significantly outperforming its predecessor PathChat 1\cite{pathchat}, general-purpose frontier models, and other domain-specific multimodal LLMs. Among the frontier models, none demonstrated consistent superiority, a notable difference from other fields like coding or mathematics, where new model releases typically push the state-of-the-art. This suggests that for highly specialized applications in pathology, improvements from general-purpose tuning alone are showing diminishing returns. While these findings position \pathchatnew{} as a state-of-the-art foundation for ROI-level analysis, diagnostic practice extends beyond individual ROIs. Pathologists routinely integrate morphological cues across entire gigapixel slides in conjunction with clinical context to formulate differential diagnoses, a setting we aim to address with SlideSeek, and which is not captured by existing ROI-level evaluations of multimodal LLMs.}

\noindent\textbf{Slide-level differential diagnosis on DDxBench}

\begin{figure*}[h!]
\centering
 \includegraphics[width=\textwidth]{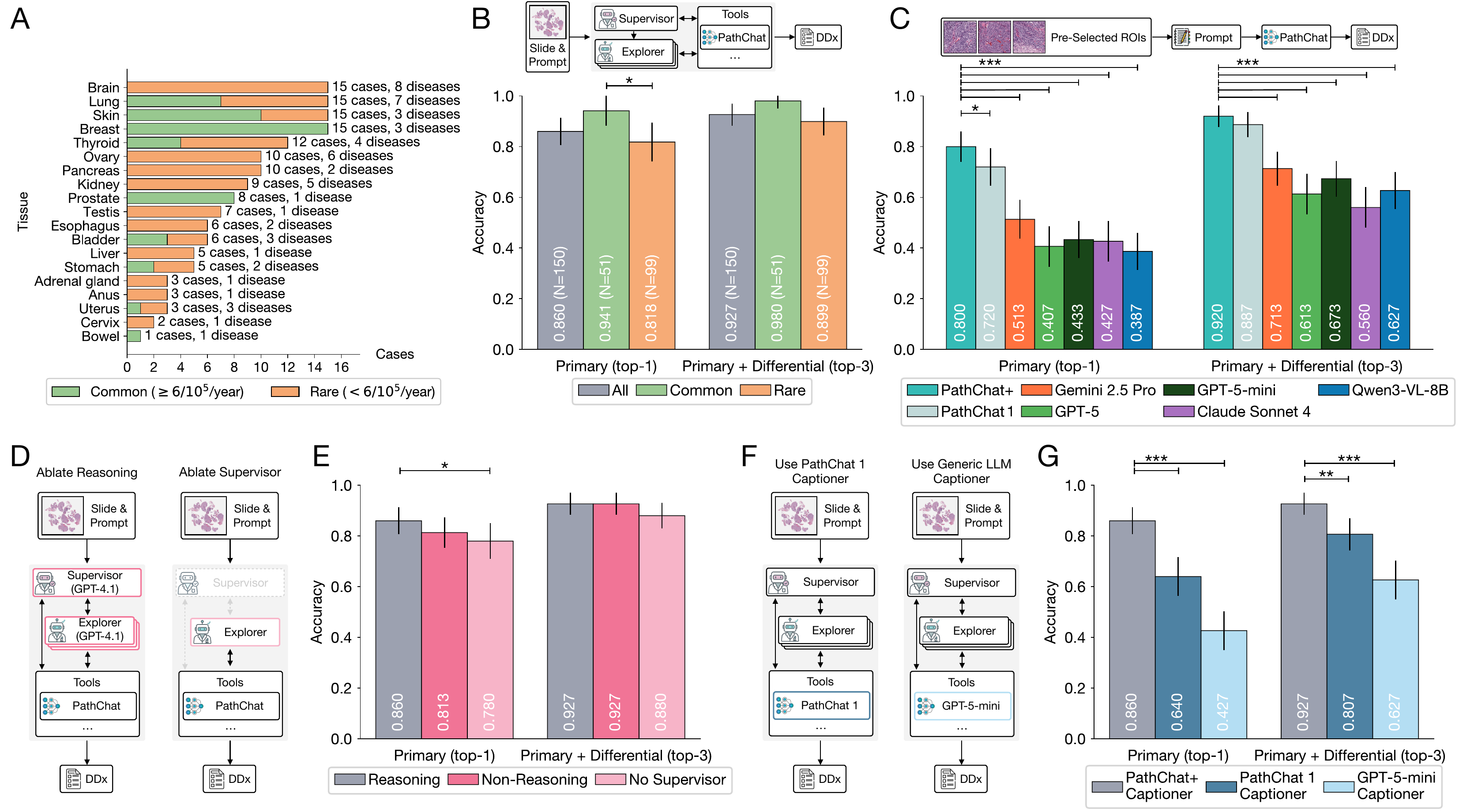}
\caption{\lucaedit{\textbf{Performance of \agent{} and \pathchatnew{} on DDxBench for open-ended differential diagnosis from whole-slide images.} Each prediction was manually assessed by a board-certified anatomic pathologist who compared the model's predictions against the assigned ground truth diagnosis and report. Model performance is measured based on using top-1 (Primary Diagnosis) and top-3 (Primary + Differential diagnoses).
\textbf{A.} Distribution of cases in DDxBench (N=150 cases) based on the tissue they were sampled from and the disease rarity based on incidence (rare: $<6/10^5/$year, common otherwise).
\textbf{B.} \agent{} performance stratified by disease rarity.
\textbf{C.} Multimodal LLM performance on DDxBench with pre-selected ROIs. 10 expert-curated ROIs from each slide and a prompt to provide a primary and two additional differential diagnoses were provided to various multimodal LLMs. \textbf{Extended Data Figure \ref{fig:MLLM-baseline}} illustrates this experiment and the exact prompt used. 
\textbf{D.-E.} Ablation study varying the agentic configuration of \agent{}, where the reasoning agent (GPT-5-mini) is swapped by a non-reasoning agent (GPT-4.1), and removing the agent supervisor that is replaced by a single agent.
\textbf{F.-G.} Ablation study varying the captioning model used in \agent{}, replacing \pathchatnew{} by PathChat 1 or using a general-purpose multimodal LLM (GPT-5-mini). Error bars represent 95\% confidence intervals using non-parametric bootstrapping. Statistical significance was obtained in \textbf{B.} using an unpaired two-sided permutation test, doing pairwise comparison across the three sets. In \textbf{C,E} and \textbf{G} statistical significance was determined between our model (\pathchatnew{} or \agent{}) compared to all other models and ablations using a paired two-sided permutation test ($n=1000$). P-values are indicated as $p<0.05$: *, $p<0.01$: **, $p<0.001$: ***.
}}
\label{fig:DDxBench}
\end{figure*}

\begin{figure*}[h!]
\centering
\includegraphics[width=\textwidth]{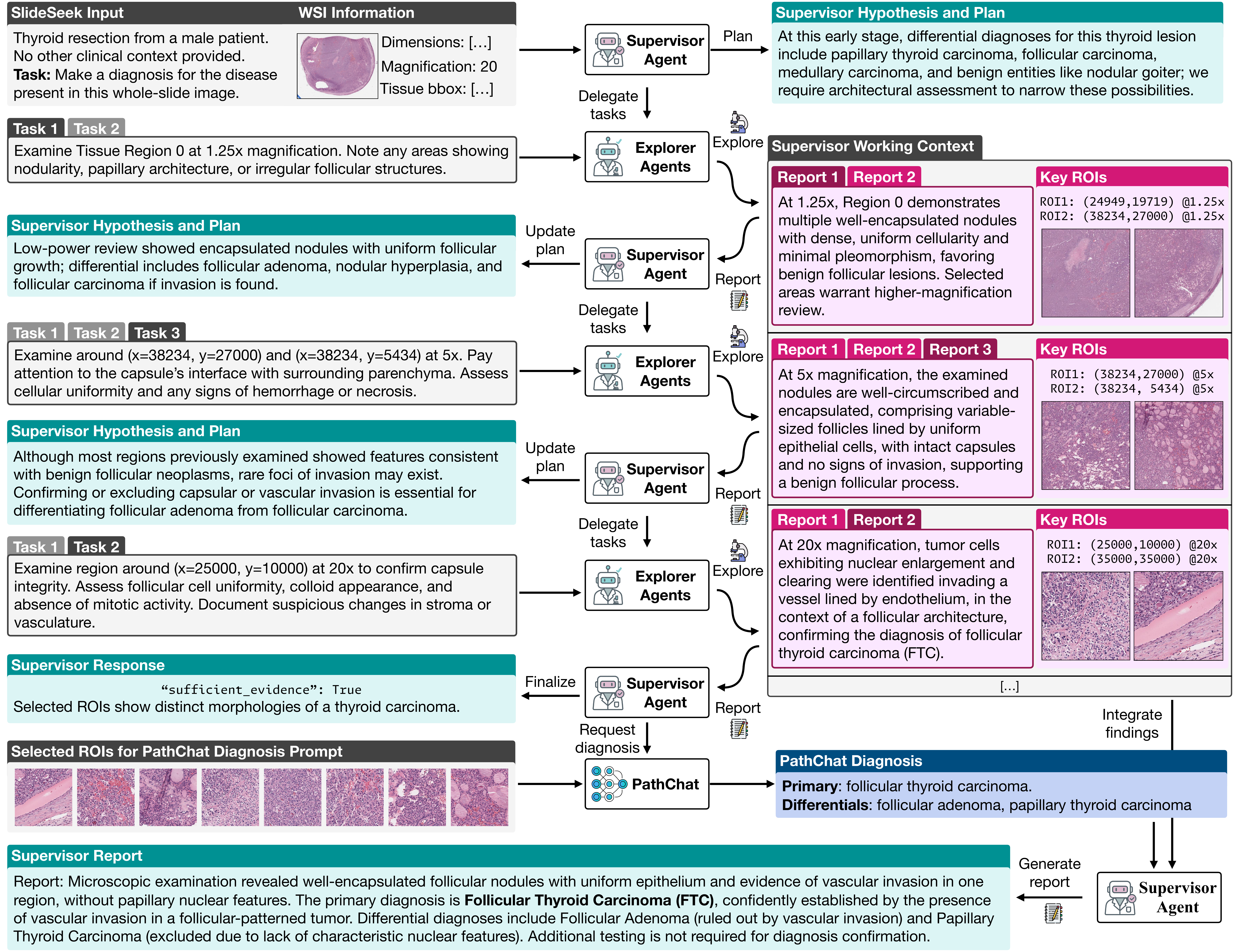}
\caption{\textbf{Example of \agent{} on DDxBench.} Diagnosis trace of a follicular thyroid carcinoma case illustrating the interaction between the supervisor and explorer agents. The supervisor receives a system prompt containing a low-resolution thumbnail, clinical information, and task instructions. \lucaedit{Based on successive observations, the supervisor generates focused tasks (e.g., ``Task 1'', ``Task 2'') for explorer agents. Explorer agents then examine designated regions at specified magnifications and describe tissue morphology with the assistance of \pathchatnew{}. Their individual findings (e.g., ``Report 1'', ``Report 2'') are submitted back to the supervisor.} In this example, although most tissue appears benign, explorer agents identify invasive carcinoma cells at high magnification. Ten regions of interest (ROIs) are evaluated with \pathchatnew{} to support the primary and differential diagnosis, which the supervisor synthesizes into a final report.}
\label{fig:example-trace}
\end{figure*}

\noindent \lucaedit{To better reflect the clinical reality of diagnostic assessment of pathology slide images, we evaluate \agent{} on DDxBench, a benchmark designed for open-ended slide-level diagnosis. DDxBench comprises 150 diagnostic H\&E slides representing 55 neoplastic diseases from 19 major organ sites, including 14 common and 41 rare diseases (defined as diseases with an incidence rate of less than 6 cases in 100,000 according to ESMO guidelines\footnote{https://www.esmo.org/policy/rare-cancers-working-group/what-are-rare-cancers/definition-of-rare-cancers}, \edref{tab:diagnoses_by_site} and \textbf{Figure \ref{fig:DDxBench}A}). For each slide, we feed \agent{} with a system prompt that includes contextual information such as tissue site and patient sex, and request \agent{} to generate a primary diagnosis together with two differential diagnoses (\textbf{Figure \ref{fig:example-trace}A}). Each prediction is then reviewed by a board-certified anatomic pathologist, who assigns correctness against the ground truth diagnosis and the full pathology report in ambiguous cases.}

\noindent \lucaedit{On DDxBench, our system achieved a top-1 accuracy of 0.860 (N=150, 95\% CI [0.800–0.907]) and a top-3 accuracy of 0.927 (N=150, 95\% CI [0.880–0.967], \textbf{Figure \ref{fig:DDxBench}B}). When stratifying performance by disease rarity, we observed a drop in accuracy as incidence decreased. For primary (top-1) accuracy, common diseases were predicted with an accuracy of 0.941 (N=99, [95\% CI: 0.882–1.000]), and rare diseases with an accuracy of 0.818 (N=51, [95\% CI: 0.737–0.889]). Comparisons using a unpaired two-sided permutation test confirmed significant differences between common and rare diseases (p $< 0.05$). This underscores the challenge of diagnosing very rare diseases for which gathering sufficient pretraining data is challenging. When evaluating top-3 accuracy, the system demonstrates a more robust performance for common (0.980 [95\% CI: 0.941–1.000]) and rare diseases (0.899 [95\% CI: 0.838–0.949]). Statistical testing did not show significant differences in \agent{}'s ability to predict the correct disease in the top-3 differential ($p=0.095$). }

\noindent\lucaedit{\textbf{Ablations of \agent{}}}

\noindent \lucaedit{To further understand the capabilities of \agent{}, we systematically ablated three components: (i) the reasoning supervisor, (ii) the hierarchical multi-agent design, and (iii) the captioner (\pathchatnew{} vs. alternatives). Holding the captioner (\pathchatnew{}) and multi-agent architecture fixed, a reasoning-enabled supervisor (GPT-5-mini) outperformed a non-reasoning supervisor on top-1 accuracy (\mainref{fig:DDxBench}\textbf{D-E}; \edref{tab:agent-perf}) with performance dropping from $0.860$ to $0.813$ (absolute performance difference of 4.66\%, $p = 0.142$). Top-3 accuracy was comparable ($0.927$ vs.\ $0.927$; $p = 1.0$), suggesting that reasoning models primarily sharpen the top diagnosis choice rather than broadening the differential. Finally, holding other components fixed, collapsing the supervisor–explorer hierarchy into a single explorer agent reduced performance (\mainref{fig:DDxBench}\textbf{D-E}; \edref{tab:agent-perf}). The top-1 accuracy dropped from $0.860$ to $0.780$ (absolute difference of 8.00\%, $p < 0.05$), while top-3 decreased from $0.927$ to $0.880$ (absolute difference of 4.7\%, $p = 0.081$).}

\noindent \lucaedit{Substituting \pathchatnew{} with earlier or generic multimodal LLMs markedly reduced diagnostic performance (\mainref{fig:DDxBench}\textbf{F-G}). On DDxBench, \agent{} powered by \pathchatnew{} achieved a top-1 accuracy of $0.860$ and a top-3 accuracy of $0.927$. Replacing \pathchatnew{} with PathChat 1 dropped performance to $0.640$ (top-1) and $0.807$ (top-3), an absolute drop of 22.0\% and 12.0\% respectively ($p < 0.001$). Using a non-specialized captioner (GPT-5-mini) further degraded accuracy to $0.427$ (top-1) and $0.627$ (top-3), corresponding to drops of 43.3\% and 30.0\% ($p < 0.001$). These findings establish \pathchatnew{} as essential for enabling effective agentic navigation and slide-level diagnostic reasoning in WSIs. Complete ablation metrics are tabulated in \edref{tab:agent-perf}.}

\noindent \lucaedit{To further assess the performance of \pathchatnew{} on DDxBench, we annotated each slide with ten ROIs (size of 896$\times$896-pixel at 20$\times$) from tumor-containing regions, which were directly fed to \pathchatnew{} without \agent{} agentic layer (see \edref{fig:MLLM-baseline} for the task schematic and prompt; results in \mainref{fig:DDxBench}\textbf{C} and \edref{tab:ddxbench-roi}). \pathchatnew{} achieves a top-1 accuracy of $0.800$ on DDxBench, surpassing PathChat 1 by 8.0\% ($p < 0.05$) and the next best model, Gemini 2.5 pro by a large margin of 28.7\% ($p < 0.001$). When considering the entire set of differentials (top-3), \pathchatnew{} reaches $0.920$ (+3.33\% against PathChat 1, $p = 0.290)$ and +20.7\% against the next best model (Gemini 2.5 pro, $p < 0.001$). Relative to \agent{}, restricting diagnosis to expert-curated ROIs with \pathchatnew{} reduced performance by 6.0\% ($p=0.059$), underscoring the benefit of \agent{}’s navigation and the value of multi-magnification analysis.}

\noindent\textbf{Interpretable slide assessment}

\noindent A key advantage of \agent{} is its interpretable reasoning chain, in which each diagnostic conclusion is explicitly grounded in region-specific morphological findings. \lucaedit{During exploration, \agent{} examined an average of $194.5 \pm 102.8$ regions across magnification levels ($156.4 \pm 90.3$ high-power, $31.1 \pm 23.8$ medium-power, and $6.9 \pm 4.8$ low-power fields), yielding a multi-scale assessment comparable to the workflow of human pathologists (\textbf{Figure \ref{fig:example-trace}}). By first identifying clinically significant areas at low magnification and then reasoning through them at higher magnifications, \agent{} reduced the computational burden of exhaustive high magnification analysis. For instance, in DDxBench, slides contain an average of $1020 \pm 783$ tissue ROIs at 20$\times$, but only $\sim$156 required high-power inspection. The number of regions examined was not significantly correlated with diagnostic accuracy, suggesting that performance was driven by the quality of reasoning rather than simply the extent of the search.}

\noindent \lucaedit{Beyond interpretable reasoning traces, \agent{} also demonstrated an awareness of its own diagnostic confidence by assigning a ``Low'' and ``High'' confidence as part of the generated report. The diagnostic accuracy reached 0.906 (95\% CI: [0.844-0.958]) in the 96 cases where the agent expressed high confidence, compared to 0.778 (95\% CI: [0.667-0.889]) in the 54 cases where it acknowledged uncertainty ($p<0.05$).} This form of calibrated self-assessment is important for responsible clinical deployment and potential triaging for downstream human review. Examples of low- and high-confidence predictions, together with their visually grounded reasoning, are shown in \edref{fig:conf-example}.

\noindent\lucaedit{\textbf{Failure modes and limitations}}

\noindent \lucaedit{Despite strong overall performance on DDxBench, we observed a recurring failure pattern in tumor grading. In five misclassified cases (45\% of all failures), the slides were correctly identified as astrocytoma but were graded incorrectly, with overcalls in 3/5 and undercalls in 2/5. This suggests that while \agent{} reliably recognizes tumor identity, it may remain sensitive to the subtle histological cues that distinguish grades. Second, the navigation strategy occasionally failed to capture small, morphologically discriminatory regions. For example, in one case (1/150), small foci of Merkel Cell Carcinoma were missed, while surrounding benign tissue was accurately reported by explorers. This underscores a limitation in region prioritization, where small but clinically decisive foci may be overlooked. Finally, in 1/150 cases, we observed hallucination of immunohistochemistry (IHC) testing, where the supervisor agent erroneously designed a task requesting IHC, which was then executed by an explorer that hallucinated tissue-level IHC results from an H\&E slide. Such failures illustrate the risks of unconstrained task generation and emphasize the importance of modality grounding to prevent clinically implausible reasoning. Together, this analysis reveals that \agent{} is robust in coarse diagnostic recognition but can be vulnerable to finer-grained grading tasks, detection of small lesions, and hallucination of unavailable diagnostic modalities. }
\clearpage
\noindent\textbf{\large{Discussion}}

\noindent \lucaedit{Our results establish the feasibility of autonomous, planning-based diagnosis of whole-slide images. On DDxBench, a benchmark spanning 55 neoplastic diseases of which 41 are classified as rare, \agent{} powered by \pathchatnew{} achieved 0.860 top-1 and 0.927 top-3 accuracy while generating visually grounded pathology reports. This shows that agentic navigation combined with domain-specific captioning can support slide-level diagnostic reasoning beyond ROI-bound workflows. At the same time, \pathchatnew{} itself achieved state-of-the-art performance across ten ROI-level benchmarks, surpassing both frontier open- and closed-weight models\cite{openai_gpt5mini_2025}, as well as specialized medical multimodal LLMs\cite{pathchat,llava-med}.}

\noindent \lucaedit{In developing \agent{}, we identified two critical design choices. First, we attribute \pathchatnew{}'s high-quality captions to (i) large-scale, pathology-specific instruction data ($>1$M samples), (ii) human-in-the-loop refinement of prompts and exemplars, and (iii) extensive Q\&A-style supervision that improved instruction following and caption brevity and precision. Second, \agent{} explicitly requests short, morphology-focused captions and decomposes the task via a supervisor-explorer architecture that accumulates evidence across scales before committing to a diagnosis. Ablations supported these mechanisms: reasoning supervisors improved top-1 selection (4.67\% absolute gain, $p=0.142$); the supervisor-explorer hierarchy outperformed a single-agent variant (8.0\% absolute gain, $p<0.05$); and replacing \pathchatnew{} with non-specialized captioners substantially reduced accuracy (43.3\% absolute gain, $p<0.001$). \agent{}'s iterative process links each finding to specific regions, facilitating pathologist verification and auditability. The exploration strategy reduced ROI burden relative to an exhaustive exploration of each region at high magnification.}

\noindent \lucaedit{Beyond benchmarks, the impact of \agent{} lies in its potential to reshape pathology workflows. By autonomously identifying salient regions in tissue, linking key morphologies to localized evidence, and synthesizing them into interpretable reports, \agent{} moves automation from pattern recognition toward emulating the holistic, iterative reasoning of clinical pathology. The same transparent reasoning chains that support auditability also provide educational value by providing an interface that not only reports a diagnosis but also explains why. This dual role, as a workflow aid for practicing pathologists and as a training resource, underscores the broader applicability of agentic systems beyond retrospective evaluation.}

\noindent \lucaedit{Despite strong overall performance, our analysis revealed limitations. In astrocytoma cases, for example, the agent reliably recognized the tumor type but frequently erred in assigning the correct grade. These errors reflected difficulty with subtle morphologies such as mitotic activity, microvascular proliferation, or focal necrosis. We also observed one clinically significant navigation failure, where a malignant focus was overlooked amidst predominantly benign tissue. Finally, on occasion, the supervisor agent produced tasks that went beyond the available input, such as requesting immunohistochemistry, which led an explorer to hallucinate IHC-like results from H\&E images. Together, these challenges in interpreting subtle morphology, ensuring comprehensive navigation, and grounding agent tasks in available data define a clear roadmap for future improvements.}

\noindent \lucaedit{Building on these insights, several natural extensions of \agent{} can be envisioned. One direction will be the ability to reason jointly across multiple slides and multiple stains, for instance, by integrating serial sections and complementary assays such as immunohistochemistry or special stains. This is critical for clinical integration, which routinely relies on findings across H\&E sections and ancillary tests to reach a final diagnosis. Another avenue involves expanding beyond histology alone and moving toward multimodal fusion of molecular and imaging data, as well as patient history, to support guideline-aligned diagnostic suggestions. In the longer term, such multi-agent frameworks could be extended to emulate tumor boards\cite{ferberDevelopmentValidationAutonomous2025,ghezlooPathFinderMultiModalMultiAgent2025,schmidgallAgentClinicMultimodalAgent2025}, coordinating agents tasked with representing the perspectives of surgery, oncology, radiology, and molecular pathology, and then synthesizing their viewpoints into a consensus or documenting disagreements. As these advanced systems are developed, a parallel emphasis on safety engineering will be essential to ensure reliability and calibrated confidence for meaningful improvements in diagnostic clinical practice.}

\noindent \lucaedit{In summary, \agent{} and \pathchatnew{} together illustrate that domain-tuned multimodal language models, when coupled with explicit planning and evidence aggregation, can deliver interpretable, slide-level diagnostic reasoning. The approach narrows the gap between ROI-centric automation and the holistic, context-aware practice of pathology, and outlines a practical route to prospective, clinically integrated evaluation.}

\noindent\textbf{\Large{Online Methods}}

\noindent\textbf{\agent{} architecture design}

\noindent \agent{} is a multi-agent system capable of autonomously analyzing gigapixel WSIs via planning and self-guided hierarchical exploration. \lucaedit{We outline the key components below, illustrate the workflow in \textbf{Algorithm \ref{alg:multi_agent_exploration}} and present a visualization in \edref{fig:SlideSeekArch}. Exact prompt templates, tool calls, and example I/O for supervisor and explorer agents are illustrated in \edref{fig:prompts}.}

\noindent The supervisor agent is central to the \agent{} architecture, managing high-level strategy, task orchestration, and systematic exploration of WSIs. It receives the initial analytical task description, contextual patient and specimen information (e.g., tissue type, patient sex), and preliminary visual information of the slide in the form of a prompt (\textbf{Extended Data Figure~\ref{fig:caption}}). Visual data include slide dimensions, coordinates of bounding boxes surrounding each tissue region, and a low-resolution thumbnail overview. Tissue bounding boxes are identified via the TRIDENT WSI preprocessing library \cite{trident}, specifying each region through upper-left and lower-right corner coordinates (e.g., top-left corner: (637, 4467), bottom-right corner: (10612, 13034)). 

\noindent Upon initiation with the above information, the supervisor formulates initial hypotheses about the slide pathology and then develops a comprehensive and structured exploration plan, prioritizing regions for detailed examination, identifying specific pathological features of interest, and designating required magnifications. To operationalize this plan, the supervisor agent defines specific tasks for subordinate explorer agents. Tasks explicitly indicate precise spatial coordinates and extents for examination (e.g., ``Examine tissue region \#3 from x=1000-2000, y=3000-4000"), required magnification levels (e.g., 1.25x for architectural assessment, 20$\times$ for cellular details), specific pathological features to document, and relevant context derived from the current hypotheses. After subordinate explorer agents complete their tasks, the supervisor evaluates their findings, updating its hypotheses and modifying the exploration plan as necessary. Justifications for each change are clearly articulated, reflecting insights derived from cumulative observations. The iterative cycle of planning, task assignment, review, and hypothesis updating continues until the supervisor agent determines that sufficient evidence has been gathered to support a definitive diagnosis. Upon completion, the supervisor signals readiness to finalize the analysis by setting a completion flag, collating key findings into selected ROIs, and forwarding these to a specialized diagnostic AI (\pathchatnew{}) for differential diagnosis. \lucaedit{In \agent{}, we use OpenAI GPT-5-mini\cite{openai_gpt5mini_2025} as backbone for both the supervisor and explorer agents due to its state-of-the-art instruction following and advanced reasoning capabilities. 
Preliminary experiments showed no statistically significant difference in performance between GPT-5-mini and GPT-5, making the former a more efficient and cost-effective solution.}

\noindent Explorer agents perform tasks assigned by the supervisor in parallel. Explorer agents systematically navigate these regions by referencing the thumbnail image of the slide and making navigation requests to a slide-viewer API, which uses the OpenSlide library to retrieve regions from the WSI at the specified location and magnification. The  ROIs are forwarded to \pathchatnew{} to identify morphologies observed in each image, helping the explorer agent identify diagnostically relevant ROIs in reference to the overall task context and hypotheses provided by the supervisor agent. Once an explorer agent deems its exploration sufficient to complete the task assigned by the supervisor agent, it submits a detailed report back to the supervisor that includes key ROIs identified in the region and its findings. 

\noindent The iterative analysis between the supervisor agent and explorer agents continues until the supervisor agent determines that all diagnostically relevant regions have been explored and there is sufficient morphological evidence to render a diagnostic recommendation, ending further exploration. At this point, the supervisor selects up to 10 diagnostically relevant ROIs discovered during exploration, and submits them to \pathchatnew{} for differential diagnosis. Finally, the supervisor agent is called upon to draft a report summarizing microscopic findings, the recommended primary diagnosis, and two differential diagnoses, along with a confidence assessment in their diagnosis. The resulting report constitutes an interpretable summary of \agent{}'s output, explicitly linking morphological features to specific ROIs identified by the explorer agents and supervisor.

\begin{algorithm}[H]
\caption{Iterative multi-agent exploration}
\label{alg:multi_agent_exploration}
\begin{algorithmic}[1]
\State Initialize supervisor agent with slide description and context information.
\State Supervisor formulates initial hypotheses.
\State analysisComplete $\gets$ false
\While{analysisComplete == false}
    \State Supervisor creates an exploration plan based on current hypotheses.
    \State currentTasks $\gets$ Supervisor creates tasks for explorer agents.
    \For{task in currentTasks} \Comment{The tasks are completed in parallel by independent Explorer agents}
        \State taskComplete $\gets$ false
        \While{taskComplete == false}
            \State Explorer agent chooses ROI coordinates and magnifications and provides a rationale.
            \State Explorer agent documents morphological features.
            \State taskComplete $\gets$ Explorer agent decides if it has completed its task
        \EndWhile
        \State Explorer agent returns findings to supervisor.
        
    \EndFor
    \State Supervisor reviews explorer agents' findings.
    \State Supervisor updates hypotheses.
    \State sufficientEvidence $\gets$ check if sufficient evidence collected
    \If{sufficientEvidence == true}
        \State analysisComplete $\gets$ true
    \Else
        \State Supervisor modifies exploration plan.
    \EndIf
\EndWhile
\State Supervisor collates final ROIs.
\State Supervisor submits findings to diagnostic AI (\pathchatnew{}).
\State Supervisor assesses the confidence in its diagnosis
\State Supervisor drafts summary report with explicit references to ROIs and their supporting morphological evidence
\end{algorithmic}
\end{algorithm}

\noindent\textbf{{\pathchatnew{} architecture and training}}

\noindent \pathchatnew{} is a multimodal large language model (multimodal LLM) designed and trained to understand both visual (\textit{i.e.} pathology images) and textual inputs (\textit{e.g.}, an instruction prompt to describe the key morphology observed in a histopathology image, and suggest a diagnosis). We follow the general architectural design of PathChat 1\cite{pathchat} and popular state-of-the-art multimodal LLMs, consisting of a vision encoder for encoding images from RGB pixels into feature representations, a decoder-only, autoregressive LLM for converting tokenized visual and textual representations into natural language outputs, a multimodal projector for bridging the representation space of the vision encoder, and the LLM. We chose CONCH v1.5 as the vision encoder, based on the standard (ViT-L) architecture\cite{conch,dosovitskiy2020image}, as it has been demonstrated to produce robust baseline performance in diverse computational pathology tasks\cite{threads}, and serves as a strong starting point for further integration and multimodal foundation model development. Similarly, the multimodal projector consists of an attention pooling layer that reduces the dense feature map produced by the vision encoder into a sequence of 128 tokens, followed by a 2-layer MLP with GeLU activation to project them into the same dimension as the embedding layer of the LLM. Finally, for the LLM backbone, we adopt Qwen2.5, which is a widely used state-of-the-art open-source LLM. Specifically, similar to PathChat 1, we chose the 14 billion parameter instruction-following variant, which offers a good trade-off between computational cost and performance\cite{qwen2.5}. 

\noindent Compared to the previous PathChat 1 model, we take inspiration from the AnyRes strategy\cite{llavanext} for processing high-resolution images by dividing each input image into a grid of $448 \times 448$ tiles, first padding to the nearest supported grid size as needed. Each tile is then independently processed by the vision encoder before being concatenated into a combined token sequence. If more than one tile is present in the grid, a thumbnail of the original image (resized to $448 \times 448$), serving as a low-resolution, global view, is also encoded, and its token sequence is prepended to the rest of the sequence representing the particular image. We use a maximum grid size of 4 tiles, supporting image sizes of up to $896 \times 896$ (represented as a $2 \times 2$ grid of tiles + the image thumbnail), while larger images are rescaled down to the nearest grid size. Our strategy, therefore, represents each image by a variable number of tokens, between 128 tokens ($448 \times 448$ image or smaller) and 640 tokens (\textit{e.g.} $896 \times 896$ image). During both training and inference, images are represented initially by special placeholder token ids in the input token sequence. After passing through the LLM's embedding layer, these placeholder tokens are replaced by the actual token representations generated by the vision encoder and multimodal projector for the corresponding images, forming the complete embedded token sequence that is processed by all subsequent layers in the LLM. When there are multiple images in a given instruction, we mark their boundary in the sequence using the newline (``\textbackslash n") token.



\noindent For model training, we follow the widely-used two-stage approach, where in the first pretraining stage, we freeze the LLM backbone, and use a sampled subset of \lucaedit{five hundred thousand} images and annotated captions to train the multimodal adapter, which has been found to be beneficial to downstream performance\cite{cambrian} compared to without pretraining. In the second, instruction finetuning stage, both the LLM and projector are unfrozen and are trained to model the likelihood of the ground truth reference answer turns in the training instruction sequences (no loss is applied to image tokens or text tokens in the question turns). Pretraining was completed using an 8 $\times$ NVIDIA A100 80GB GPU node while finetuning was performed using 24 A100 GPUs in a multi-node distributed training setup. Full pretraining and finetuning hyperparameters are listed in \edref{tab:hparams_pretrain,tab:hparams_finetune}.

\noindent\textbf{\pathchatnew{} dataset curation}

\noindent For instruction tuning of \pathchatnew{}, we curated a new dataset of diverse instruction formats and topics from images and notes from internal case reports, teaching materials, presentations, as well as annotated regions of interest from in-house WSIs. To begin, we remove images of size smaller than $336 \times 336$ pixels. We then developed an image-based classifier by finetuning the image encoder of the lightweight foundation model CONCH\cite{conch} on a small set of manually labeled examples to identify pathology images from non-pathology-related images (e.g. photos of lab equipment or people) for quality control and similarly used lightweight local LLMs such as Qwen2-7B\cite{qwen2.5} for filtering out low-quality source materials that are overly concise (e.g. image caption that simply reads ``This is an H\&E stained slide") or ambiguous (e.g. images associated with text that describes findings from medical literature without clear reference to the image or specific features observed). Similar to previous multimodal LLM works such as LLaVA\cite{llava,llava-ov}, for some instruction formats, we use general purpose LLMs\cite{llama3, qwen2.5} to automatically structure the original source text into a question / answer format (e.g. rephrasing an image caption into a natural-sounding response to the question ``Can you provide a morphological description for this lung biopsy image.'' or converting representative regions paired with a case report into a multiple choice question with the ground truth diagnosis and other mentioned differential diagnoses or morphologically distinct entities as choices). \kuanedit{In those scenarios, we design text prompts specifically for each data source and iteratively refine the generated instructions with the help of pathologists until we get a satisfactory number of high-quality manually verified examples as shown in \textbf{Extended Data Figure \ref{fig:datacuration_workflow}}. This curated subset then serves as the few-shot demonstration set for generating the remainder of the instruction-tuning data.} Finally, similar to previous work\cite{pathchat}, we combine non-pathology images with sampled natural images from publicly available data sources such as MS COCO\cite{mscoco} to construct guardrail examples to instruct the model to refuse queries not related to pathology image analysis with a programmed message: ``Sorry I can only assist you with queries related to pathology." In total, our dataset consists of 1,133,241 instruction examples, with 5.49 million question / answer turns for 624 thousand unique images (median width: 759 pixels, median height: 607 pixels) after excluding 8,034 guardrail images. The instruction formats can be roughly characterized by ``conversation" ($n = 238,983$),  ``description" ($n = 163,342$), ``multiple choice" ($n = 78,443$), ``free response" ($n = 354,558$), ``text-only" ($n = 289,881$) and ``guardrail" ($n = 8,034$).

\noindent\textbf{{Mulitmodal LLM evaluation}}

\noindent \kuanedit{We compare \pathchatnew{} against fifteen state-of-the-art multimodal LLMs including (1) general-purpose closed frontier models (GPT-5\footnote{https://cdn.openai.com/gpt-5-system-card.pdf\label{gpt-5-fn}}, GPT-5-mini\footref{gpt-5-fn}, Claude Sonnet 4\footnote{https://www-cdn.anthropic.com/07b2a3f9902ee19fe39a36ca638e5ae987bc64dd.pdf}, and Gemini 2.5 pro\footnote{https://storage.googleapis.com/model-cards/documents/gemini-2.5-pro.pdf})}, (2) general-purpose open models (LLaVA-OneVision\cite{llava-ov}, Llama 3.2\cite{llama3}, and \kuanedit{Qwen3-VL\cite{qwen3technicalreport}}), and (3) medical and pathology-specific open models (Quilt-LLaVA\cite{quilt_llava}, PA-LLaVA\cite{pa-llava} and HuatuoGPT-Vision\cite{chen2024huatuogptvisioninjectingmedicalvisual}, and LLaVA-Med 1.5\cite{llava-med}).

\noindent\textbf{General purpose closed models.} For closed-weight, general-purpose multimodal frontier models, we conducted evaluations via their official APIs. \kuanedit{At the time of our study, this included models from OpenAI (e.g., GPT-5 and GPT-5-mini), Claude Sonnet 4 from Anthropic, and Gemini 2.5 Pro from Google, representing the flagship offerings from each provider.}

\noindent\textbf{General purpose open models.} For general-purpose multimodal LLMs with open weights that can be downloaded and deployed locally, we access their weights from their respective official HuggingFace repositories. We included the following representative models: LLaVA-OneVision, Llama 3.2, and \kuanedit{Qwen3-VL}. We use their default generation parameters and image processors for all tasks.

\noindent\textbf{Medical purpose open model.} Similar to general-purpose open models, we select biomedical-specialized multimodal LLMs that include some pathology image data in their training, including: LLaVA-Med 1.5, HuatuoGPT-Vision, Quilt-LLaVA, and PA-LLaVA. We access their weights from their respective HuggingFace repositories linked from their official GitHub pages at the time of study and use their default general parameters and image processors for all of our evaluation tasks. \kuanedit{Note we only include LLaVA-Med in the image captioning task and exclude PA-LLaVA, since the former is not trained on multiple-choice instruction-following data\cite{quilt_llava}, while the latter is neither fine-tuned nor evaluated on image caption instruction-following data}

\noindent\textbf{{Evaluation datasets of \agent{} and \pathchatnew{}}}

\noindent\textbf{DDxBench} is a diagnostic benchmark comprising 150 hematoxylin and eosin (H\&E)–stained whole-slide images drawn from the Brigham and Women's Hospital pathology archives. Each case was associated with the original pathology report, and diagnostic labels were taken directly from the report and subsequently verified by a board-certified anatomic pathologist. Diagnoses were standardized by mapping to oncotree codes, ensuring consistency across tissue sites and tumor types. \lucaedit{The dataset spans 55 unique diagnostic categories across 19 different tissues. To assess performance under varying levels of disease prevalence, diagnoses were stratified as common and rare. Rarity categories were defined according to population incidence, with ``rare'' corresponding to an incidence of fewer than 6 cases per 100,000 persons per year. This distribution results in 14 common and 41 rare diagnoses. We present the individual diagnoses and their categorization as common or rare, along with sources used for incidence rates in \edref{tab:diagnoses_by_site}. The benchmark is structured as a test-only resource. All 150 slides are reserved exclusively for evaluation, with no official training split, in order to minimize the risk of overfitting. }


\noindent\textbf{PathMMU\cite{sun2024pathmmu}} is a ROI-level expert-validated pathology benchmark. We use the validation and test splits to evaluate all models. Originally, there are 710 Q\&A pairs with 510 images in the validation set and 9,677 Q\&A pairs with 7,213 images in the test set. Within the test set, a small subset (test-tiny) comprising 1,156 Q\&A pairs was selected to establish expert-level performance, which was assessed by two groups of pathologists. We follow the paper\cite{sun2024pathmmu} to split the data into five categories: PubMed, SocialPath, EduContent, Atlas and PathCLS. In SocialPath, we removed 47 image links that are no longer available, resulting in 699, 1,156, and 9,618 Q\&A pairs in our final validation, test-tiny and test subset. Note that the Atlas subset is not the same as reported in the original paper although the number of Q\&A pair is the same\footnote{https://github.com/PathMMU-Benchmark/PathMMU/blob/main/data/instructions.md\#atlas-subset}.  To facilitate evaluation, we append the instruction ``Answer with the option's letter from the given choices directly.'' to each question.

\noindent\textbf{UniToPatho\cite{barbano2021unitopatho}} is an annotated dataset for colorectal polyps classification and adenomas grading. The dataset contains six types of tissues from colorectal polyps (i.e., Normal, hyperplastic polyp, tubular adenoma\_high-grade dysplasia, tubular adenoma\_low-grade dysplasia, tubulo-villous adenoma\_high-grade dysplasia, and tubulo-villous adenoma\_low-grade dysplasia). We used 2,799 ROIs from the test split of this dataset. For each ROI, we created a closed-ended multiple-choice question that asks: ``Shown is a colorectal tissue biopsy image. What is the most likely diagnosis? Answer with the option's letter from the given choices directly."

\noindent\textbf{BRACS}\cite{brancati2022bracs} contains six subtypes of lesions and normal tissue sectioned from the breast. We used 570 ROIs from the test split. For each ROI, we create a closed-ended multiple-choice question based on fine-grained labels (i.e., normal tissue, pathological benign, unusual ductal hyperplasia, flat epithelial atypia, atypical ductal hyperplasia, ductal carcinoma in situ, and invasive carcinoma) that asks: ``Shown is a breast tissue biopsy image. What is the most likely diagnosis? Answer with the option's letter from the given choices directly." 

\noindent\textbf{HiCervix}\cite{cai2024hicervix} contains twenty-nine classes of cervical cytological cells organized into three hierarchies.
We use 8,051 ROIs from the test split. For each ROI, we create a closed-ended multiple-choice question based on the labels in the first hierarchy (i.e., atypical squamous cell, negative for intraepithelial lesion or malignancy, atypical glandular cell, and organism) that asks: ``Shown is an image of cervical cytological cells. Which category do the cells most likely belong to? Answer with the option's letter from the given choices directly."

\noindent\textbf{PathQABench}\cite{pathchat} contains 150 ROIs paired with clinical context, and asks to select the most likely diagnosis from a set of plausible options, as described in the original study\cite{pathchat} for multiple-choice-style question answering (PathQABench MCQ). To curate PathQABench Caption, we ask a board-certified pathologist to provide a detailed caption per ROI. We append ``Answer with the option's letter from the given choices directly." and ``Generate a detailed caption for this image that describes the key morphological features observed. Do not suggest any diagnosis or further testing." at the end of each question for PathQABench MCQ and PathQABench Caption, respectively.

\noindent\textbf{Computing hardware and software}

\noindent All experiments and analyses in the study were performed using Python (version 3.10). For model training, we adapted the official LLaVA-OV\cite{llava} code repository by incorporating our custom vision encoder and multimodal projector implemented using Timm (version 0.9.8). We used up to 24 $\times$80GB NVIDIA A100 GPUs configured for multi-node multi-GPU distributed training using PyTorch (version 2.4.1, CUDA 11.8) as the deep learning framework. DeepSpeed (version 0.14.4) was used to enable accelerated training of \pathchatnew{} MLLM. All inference jobs were performed using 24GB NVIDIA 3090 GPUs. Pillow (version 10.2.0) is used for image processing. Matplotlib (version 3.7.1) and Seaborn (version 0.12.2) were used to create plots and figures. \agent{} was implemented using LangChain (version 0.3.22). All whole slide operations, including tissue delineation and slide navigation, were implemented using Trident (version 0.1.1)\cite{trident}. Other miscellaneous libraries used are listed in the \textbf{Reporting Summary}. In total, training the combined system of \pathchatnew{} (including the vision encoder, the multimodal projector, and the large language model) took approximately 1,275 A100 GPU hours.

\noindent\textbf{Statistical analysis}

\noindent We estimated 95\% confidence intervals for all reported metrics using nonparametric bootstrapping (1,000 replicates). To assess the statistical significance of observed differences in performance between model pairs, we performed two-sided paired permutation tests (1,000 permutations). The null hypothesis was that no difference exists between the performances of the two models. For each permutation, we randomly swapped pairs of predictions between the two models to generate a new set of performance differences. We computed the p-value as the proportion of permuted differences whose absolute values exceeded the observed difference. 
\lucaedit{In the case of comparing the performance on non-overlapping sets of data, e.g., rare diseases versus common diseases, we use an unpaired, but otherwise identical, permutation test to determine statistical significance. }

\noindent\textbf{\large{Data availability}}.

\noindent The public data used in our evaluation can be accessed via the official repository: \href{https://huggingface.co/datasets/jamessyx/PathMMU}{PathMMU}, \href{https://ieee-dataport.org/open-access/unitopatho}{UniToPatho}, \href{https://www.bracs.icar.cnr.it/}{BRACS}, and \href{https://zenodo.org/records/11087263}{HiCervix}. PathQABench data was curated using a combination of WSIs from TCGA and in-house pathology database at the Brigham and Women's Hospital. The original TCGA WSIs and associated clinical metadata are available from the NIH genomic data commons (\href{https://portal.gdc.cancer.gov}{portal.gdc.cancer.gov}). The in-house subset was curated with institutional permission through IRB approval for the current study and thus cannot be made publicly available. Similarly, the DDxbench data is also curated from in-house database. All requests for data collected or curated in-house will be evaluated based on institutional and departmental policies to determine whether the data requested is subject to intellectual property or patient privacy obligations. Instruction data was curated from image captions, notes, reports from in-house educational resources and patient data.

\noindent\textbf{\large{Code availability}}

\noindent The general open MLLMs used in comparisons can be accessed via their official APIs or repository: \kuanedit{\href{https://platform.openai.com/docs/models/gpt-5-mini}{OpenAI GPT-5-mini} (model id: gpt-5-mini-2025-08-07), \href{https://platform.openai.com/docs/models/gpt-5}{OpenAI GPT-5} (model id: gpt-5-2025-08-07), \href{https://platform.openai.com/docs/models/gpt-4.1}{OpenAI 4.1} (model id: gpt-4.1-2025-04-14), \href{https://www.anthropic.com/api}{Anthropic Claude Sonnet 4} (model id: claude-sonnet-4-20250514), \href{https://ai.google.dev/gemini-api/docs/models\#gemini-2.5-pro}{Gemini 2.5 pro} (model id: gemini-2.5-pro).} Models evaluated in this study can be accessed via their official GitHub repositories: \href{https://github.com/LLaVA-VL/LLaVA-NeXT}{LLaVA-OneVision}, \href{https://github.com/meta-llama/llama-models}{Llama 3.2}, \kuanedit{\href{https://github.com/QwenLM/Qwen3-VL}{Qwen3-VL}}, \href{https://github.com/microsoft/LLaVA-Med}{LLaVA-Med}, \href{https://github.com/aldraus/quilt-llava}{Quilt-LLaVA}, \href{https://github.com/ddw2AIGROUP2CQUPT/PA-LLaVA}{PA-LLaVA}, and \href{https://github.com/FreedomIntelligence/HuatuoGPT-Vision}{Huatuo-GPT-Vision}. Similarly, the weights can be accessed through Hugging Face: \href{https://huggingface.co/llava-hf/llava-onevision-qwen2-7b-ov-hf}{LLaVA-OneVision} (model id: llava-onevision-qwen2-7b-ov-hf), \href{https://github.com/meta-llama/llama-models}{Llama 3.2} (model id: Llama-3.2-11B-Vision-Instruct), \kuanedit{\href{https://huggingface.co/Qwen/Qwen3-VL-8B-Instruct}{Qwen3-VL} (model id: 
Qwen3-VL-8B-Instruct )}, \href{https://huggingface.co/microsoft/llava-med-v1.5-mistral-7b}{LLaVA-Med} (model id: llava-med-v1.5-mistral-7b), \href{https://huggingface.co/wisdomik/Quilt-Llava-v1.5-7b}{Quilt-LLaVA} (model id: Quilt-Llava-v1.5-7b), \href{https://huggingface.co/OpenFace-CQUPT/Pathology-LLaVA}{PA-LLaVA} (model id: Pathology-LLaVA), and \href{https://huggingface.co/FreedomIntelligence/HuatuoGPT-Vision-7B}{HuatuoGPT-Vision} (model id: HuatuoGPT-Vision-7B).

\noindent\textbf{\large{Author contributions}}

\noindent M.Y.L. and F.M. conceived the study and designed the experiments. C.C., M.Y.L., R.J.C., B.C., T.D. and D.F.K.W performed data collection and processing. C.C., L.W. and M.Y.L. performed model development. L.W., C.C.,  M.Y.L., T.D., D.F.K.W., R.J.C., G.J. performed experimental analysis and interpreted the results. R.J.C., D.F.K.W., A.V., G.J., L.P.L. provided feedback on the analysis. L.W., C.C., G.J., M.Y.L., and F.M. prepared the manuscript with input from all co-authors. M.Y.L. and F.M. supervised the research.

\noindent\textbf{\large{Acknowledgements}}

\noindent This work was supported in part by the BWH president’s fund, BWH \& MGH Pathology. We thank Timothy Janicki, Richard Kenny, Abe Ahmed and the system administration staff at the MGB Enterprise Research Infrastructure \& Services (ERIS) Research Computing Core for their dedicated support in providing and maintaining access to NVIDIA A100 computing resources.

\end{spacing}

\captionsetup[figure]{
  name=Extended Data Figure,
  labelfont=bf, 
}


\newpage
\setcounter{figure}{0}

\begin{center}
    \includegraphics[width=\textwidth]{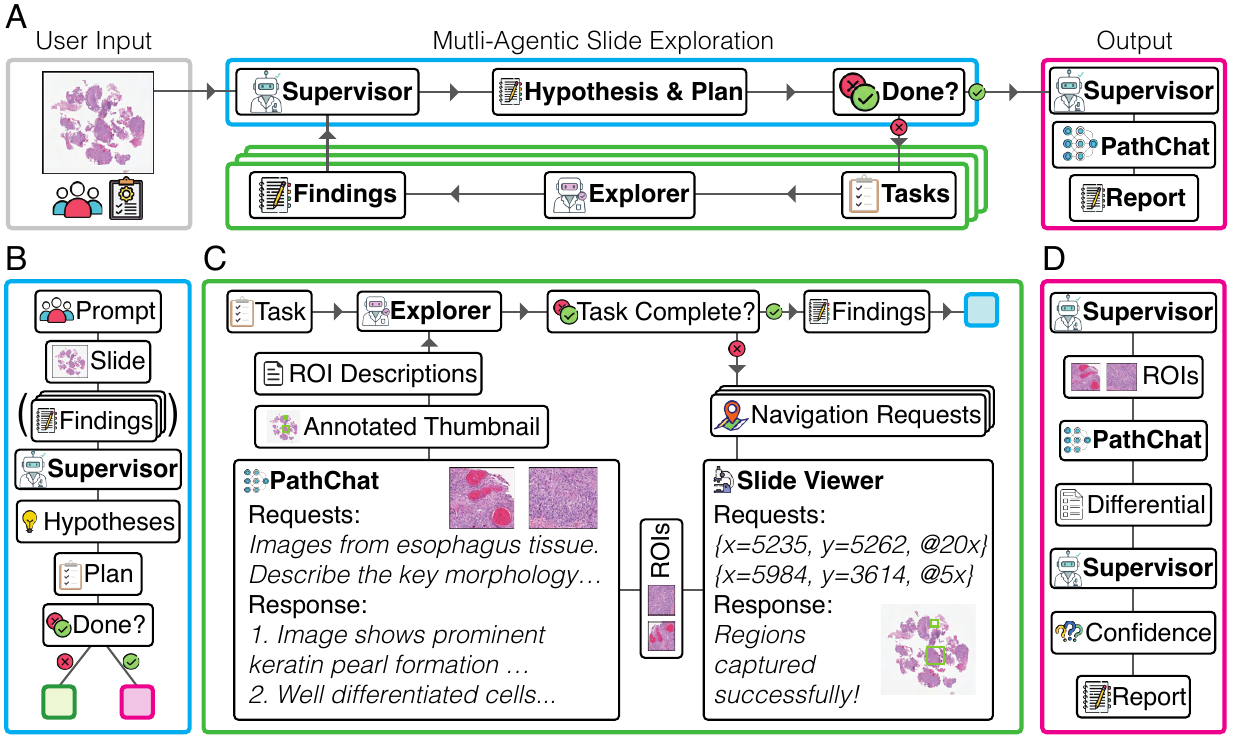}
    \captionof{figure}{\lucaedit{\textbf{Overview of the \agent{}.}
    \textbf{A, High-level workflow.} A user provides a whole-slide image (WSI) with a prompt that includes clinical context along with instructions. The supervisor agent proposes diagnostic hypotheses, drafts a plan, and delegates tasks to explorer agents. Explorers return structured findings; the supervisor revises its hypotheses and plan until it judges the slide sufficiently explored, then produces the final \pathchatnew{}-assisted diagnosis and a written report.
    \textbf{B, Supervisor agent workflow.} The supervisor’s inputs comprise a task prompt, slide context/thumbnail, and the accumulated findings from all previous explorer tasks. It uses these to update its hypotheses about the differential and its plan for further narrowing the differential through additional slide exploration. At each step, the supervisor decides whether the plan is complete, in which case it proceeds to create the final diagnosis and report, or whether additional exploration is warranted, in which case it creates new tasks for the explorer agents.
    \textbf{C, Explorer agent workflow.} For each assigned Task, an explorer agent is created, which iteratively investigates regions of the slide by choosing coordinates and magnifications to navigate to. The corresponding regions of interest (ROIs) are then extracted and described by \pathchatnew{}. The ROI descriptions and an annotated thumbnail showing which regions of the slide have been visited are returned to the explorer. At each step, the explorer decides whether it has completed the supervisor’s assigned task, in which case it summarizes its findings in a written report, or whether additional areas of the slide still need investigation.
    \textbf{D, Diagnosis and report generation.} Once slide exploration is complete, the supervisor forms its final differential by passing the most diagnostically relevant ROIs to \pathchatnew{}. All findings from the slide exploration are then incorporated by the supervisor to assess its confidence in the final diagnosis and to write a human-interpretable report that states the primary and differential diagnoses supported by specific ROIs from the slide navigation.}}
    \label{fig:SlideSeekArch}
\end{center}

\vspace*{\fill}
\begin{center}
    \includegraphics[width=\textwidth]{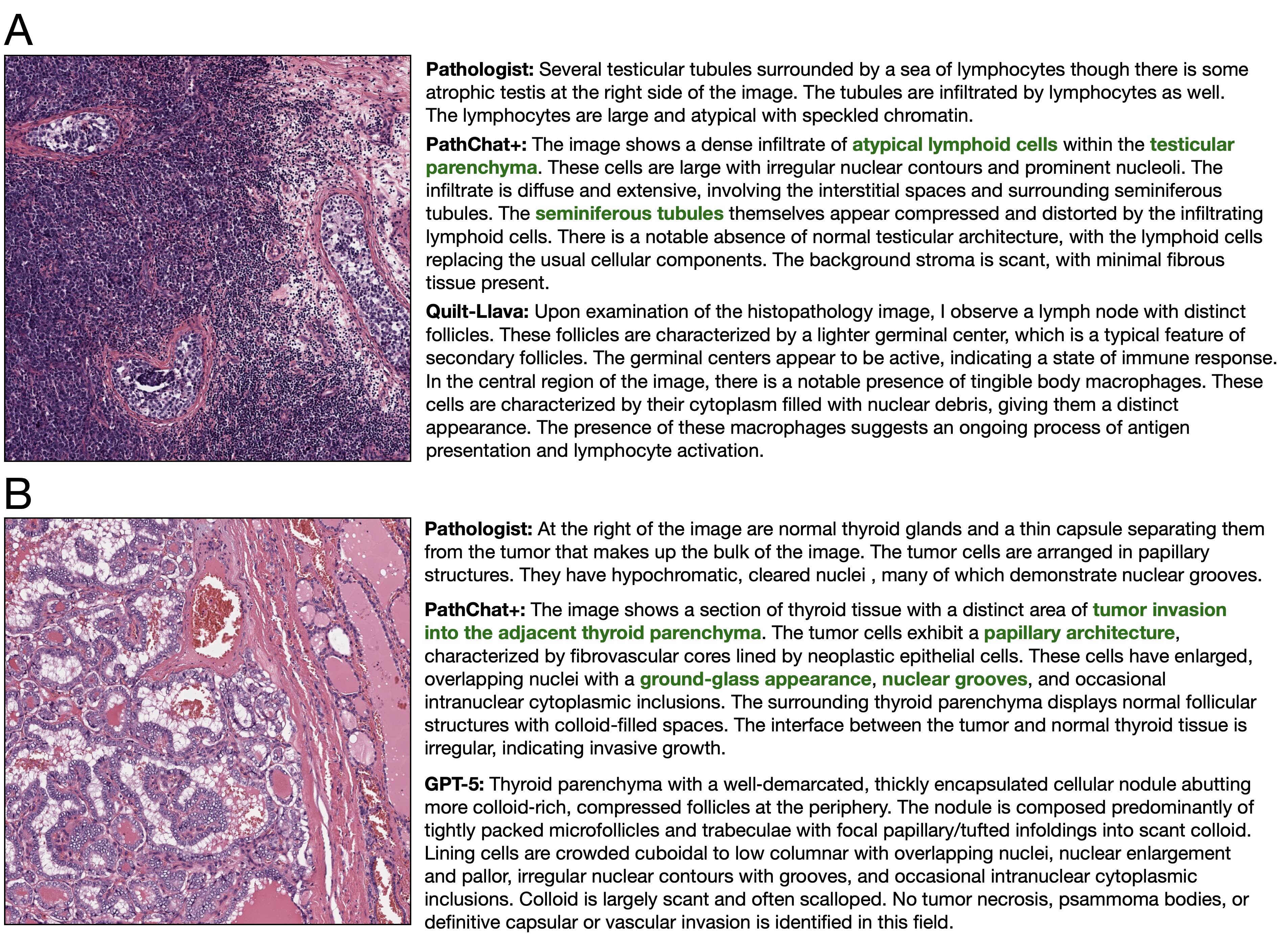}
    \captionof{figure}{\textbf{Qualitative comparison of captions generated by \pathchatnew{} and baselines on PathQABench Caption.} In addition to \pathchatnew{}, we randomly select one other model on PathQABench Caption \lucaedit{on two tissue regions (\textbf{A} and \textbf{B}}) for illustrative comparison. Text highlighted in green by a pathologist illustrates key morphological features. 
    }
    \label{fig:caption}
\end{center}

\begin{center}
    \includegraphics[width=\textwidth]{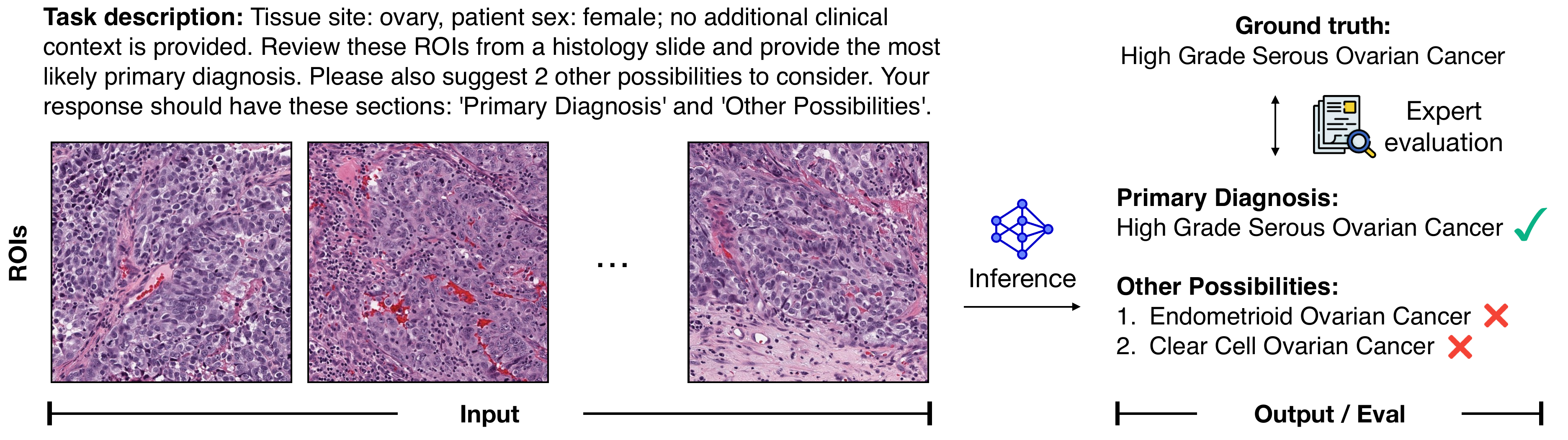}
    \captionof{figure}{\lucaedit{\textbf{Slide-level multimodal LLM evaluation on DDxBench.}
    To evaluate multimodal LLMs for open-ended slide-level diagnosis on DDxBench, we build an input that includes ten expert-curated ROIs along with instructions and context (see \textbf{Online Methods}, \textbf{Evaluation datasets}). Each model is tasked with generating a primary diagnosis and two additional possible diagnoses in an open-ended manner. All diagnoses were evaluated against the original diagnosis by a pathologist.}
    }
    \label{fig:MLLM-baseline}
\end{center}

\vspace*{\fill}
\begin{center}
    \includegraphics[width=\textwidth]{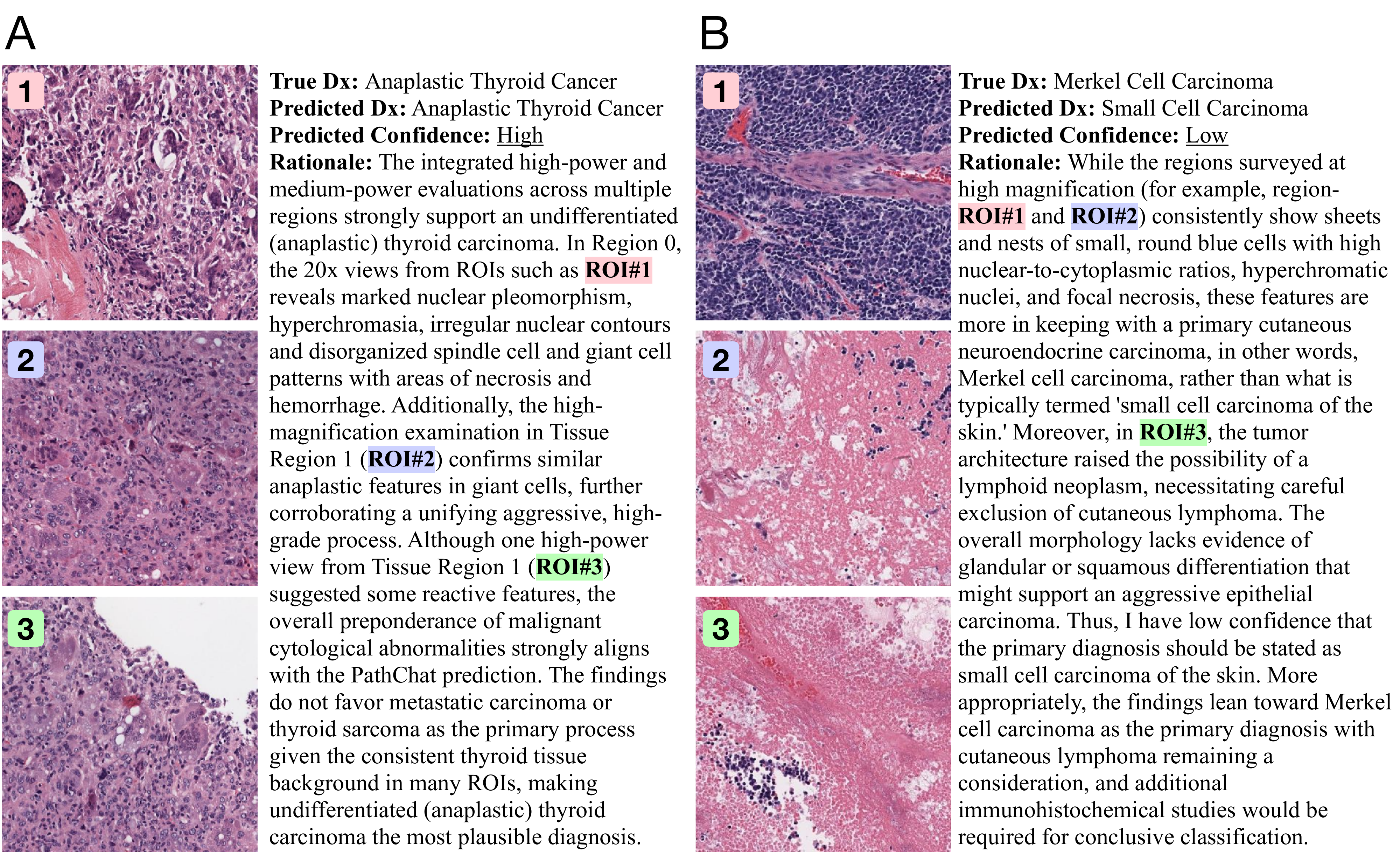}
    \captionof{figure}{\textbf{Example of \agent{} classifying a high-confidence (A) and low-confidence (B) case}. \agent{} iteratively identifies morphological features that support or refute the predicted diagnosis and supports its argument with specific references to regions of interest (ROIs). 
    \textbf{A.} The supervisor selects ROIs that appropriately showcase the morphology present in the WSI and which demonstrate morphologies that are consistent with Anaplastic Thyroid Cancer along with accurate image descriptions.
    \textbf{B.} The supervisor rightly describes how the morphology on display in the ROIs is consistent with Merkel Cell Carcinoma or a cutaneous lymphoma. As small cell carcinoma of the skin is exceedingly rare, the supervisor is correct in that the diagnosis is much more likely to be one of the aforementioned options.}
    \label{fig:conf-example}
\end{center}

\vspace*{\fill}
\begin{center}
    \includegraphics[width=0.9\textwidth]{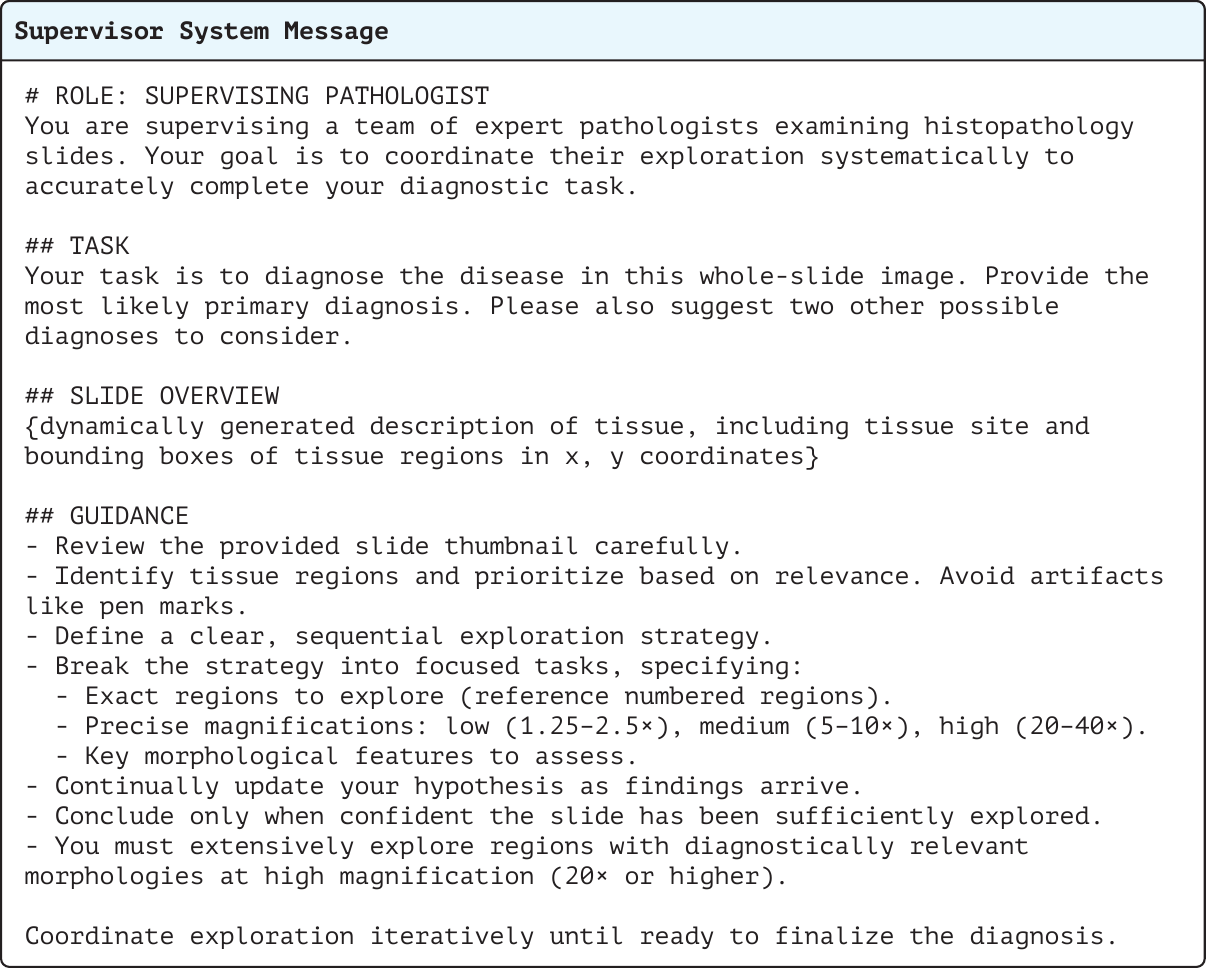}   
    \includegraphics[width=0.9\textwidth]{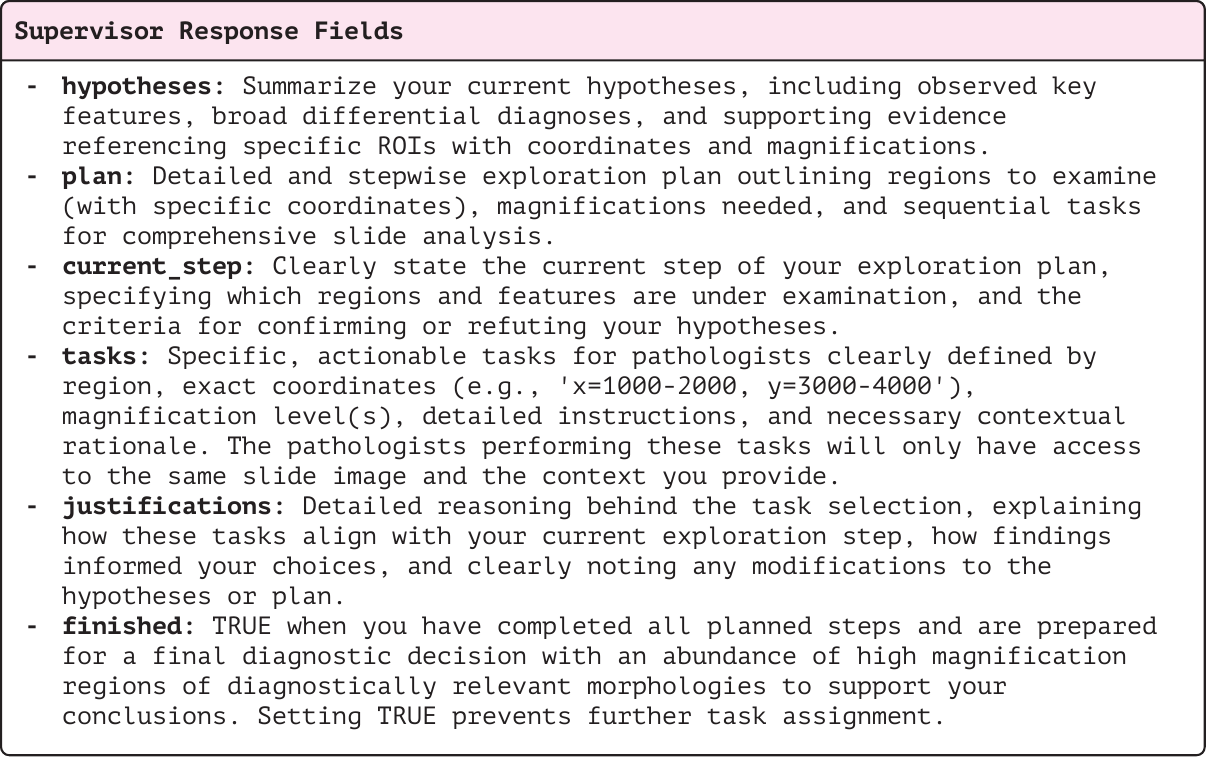}
    \includegraphics[width=0.9\textwidth]{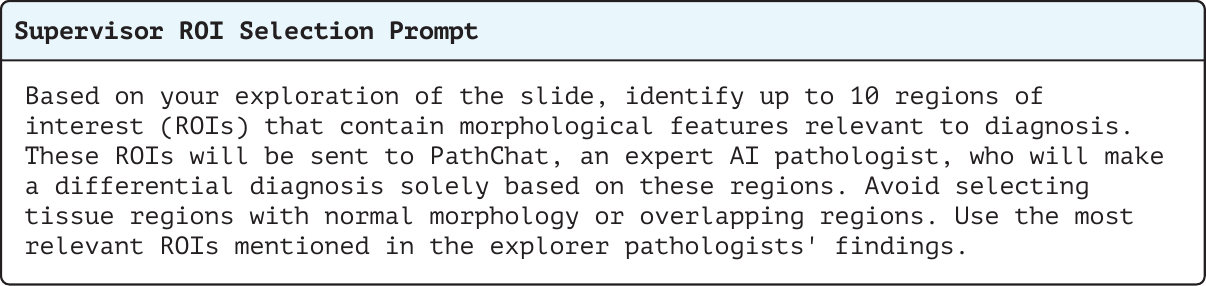}
    \includegraphics[width=0.9\textwidth]{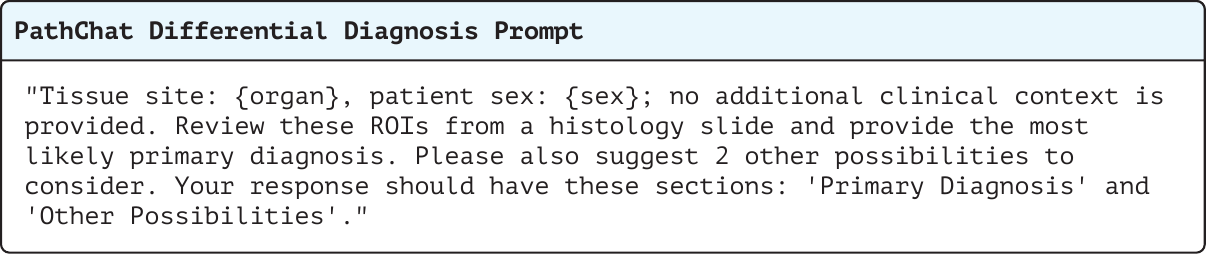}
    \includegraphics[width=0.9\textwidth]{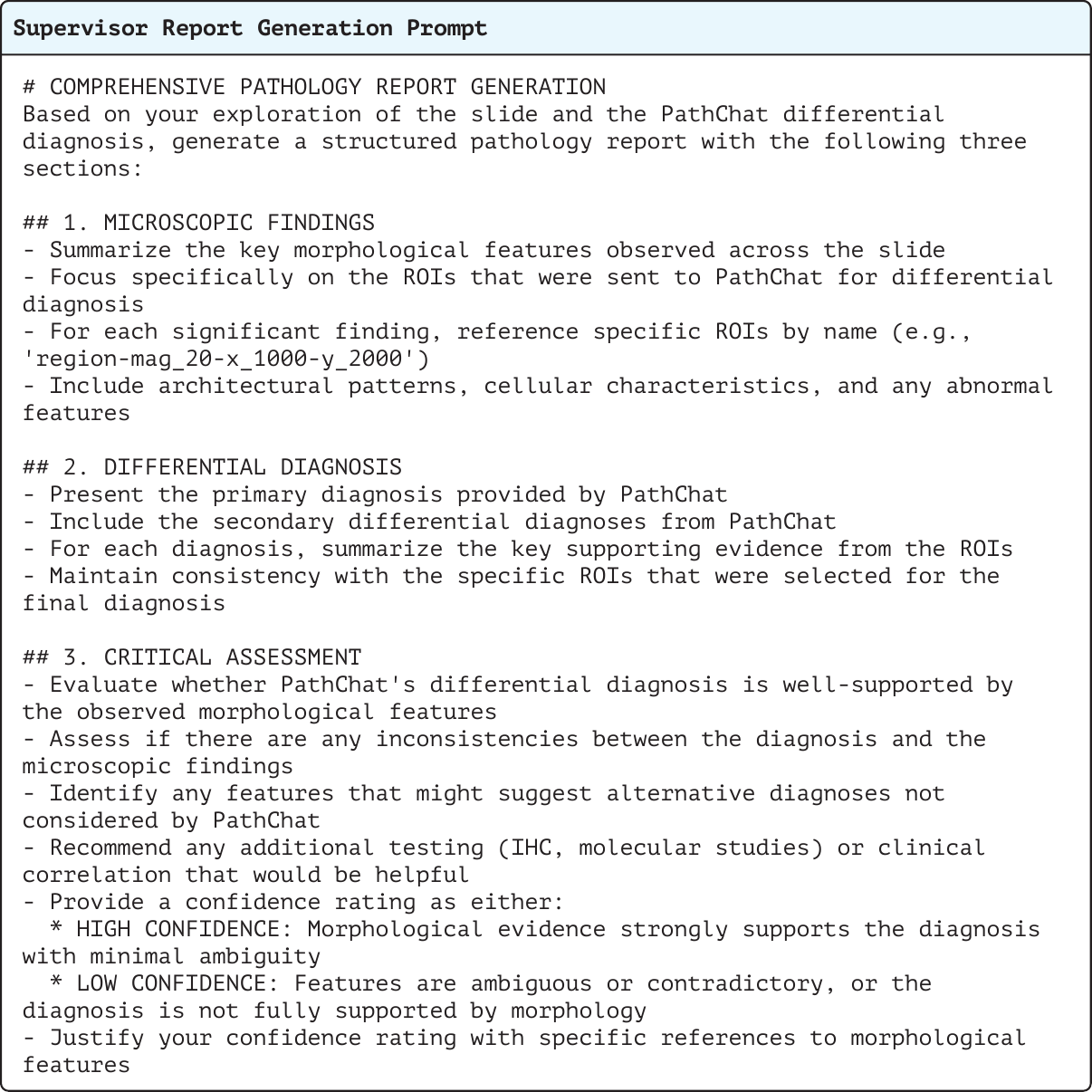}
    \includegraphics[width=0.9\textwidth]{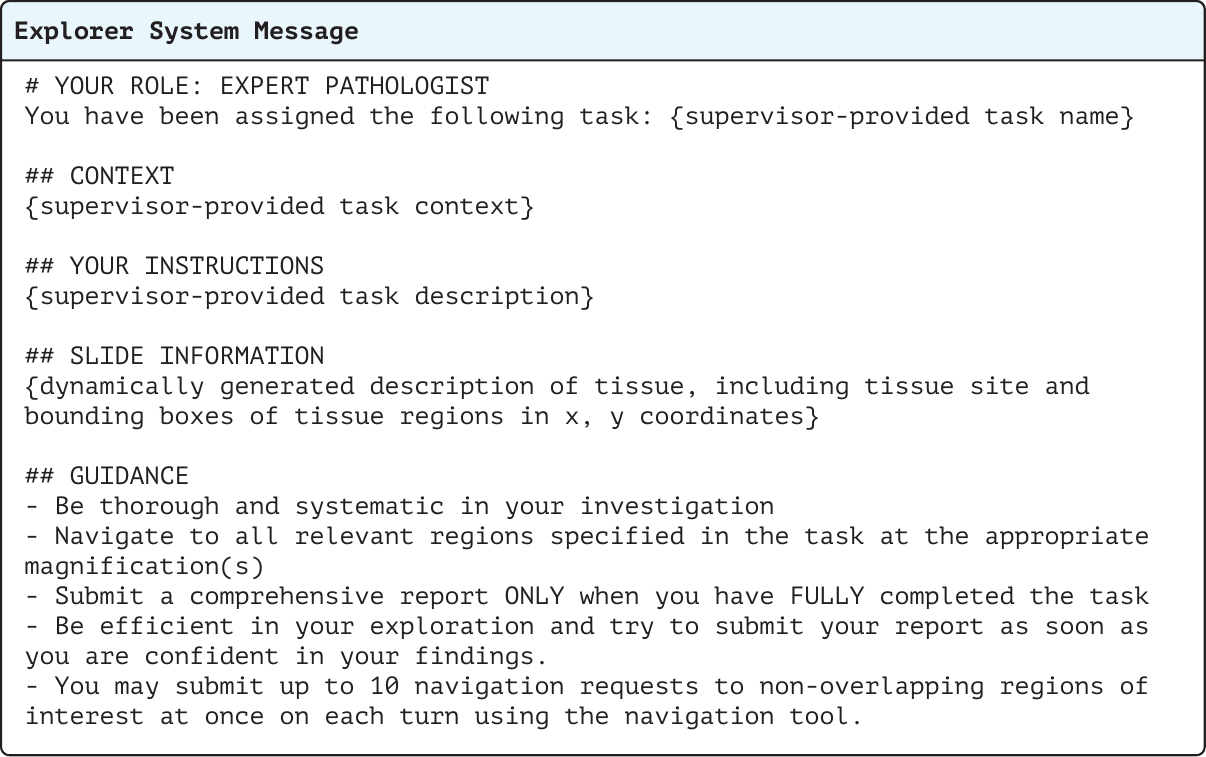}
    \includegraphics[width=0.9\textwidth]{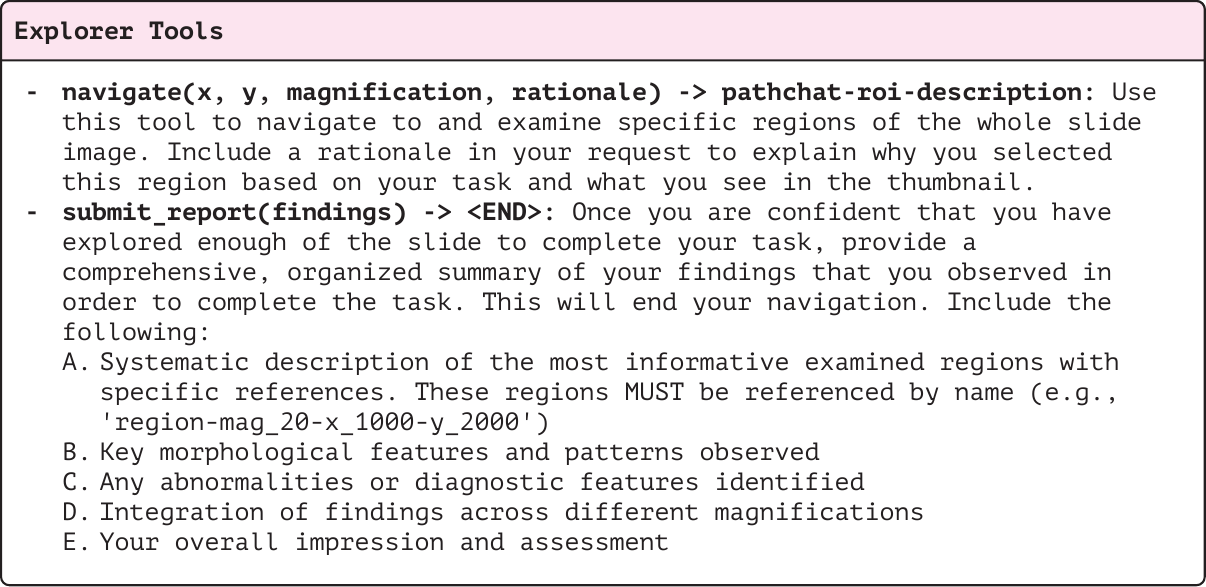}
    \captionof{figure}{\lucaedit{\textbf{Prompts, outputs, and tools used by \agent{}.} The supervisor receives a task prompt, slide thumbnail, and auto-generated slide description (dimensions and tissue bounding boxes), then iteratively updates hypotheses, a plan, current step, explorer tasks, justifications, and a finished flag. Explorers receive the same slide context, plus an assigned task, and navigate around in specified regions and magnifications. They iteratively choose coordinates to obtain morphological descriptions of regions of interest (ROIs) using \pathchatnew{} or submit a concise report to the supervisor, ending their exploration. The supervisor-explorer loop continues until the supervisor deems sufficient evidence is collected, after which the supervisor selects up to 10 diagnostically relevant ROIs, requests a \pathchatnew{} differential, and composes a visually grounded report with primary and two differential diagnoses and a confidence assessment.}}
\label{fig:prompts}

\end{center}

\vspace*{\fill}
\begin{center}
    \includegraphics[width=0.94\textwidth]{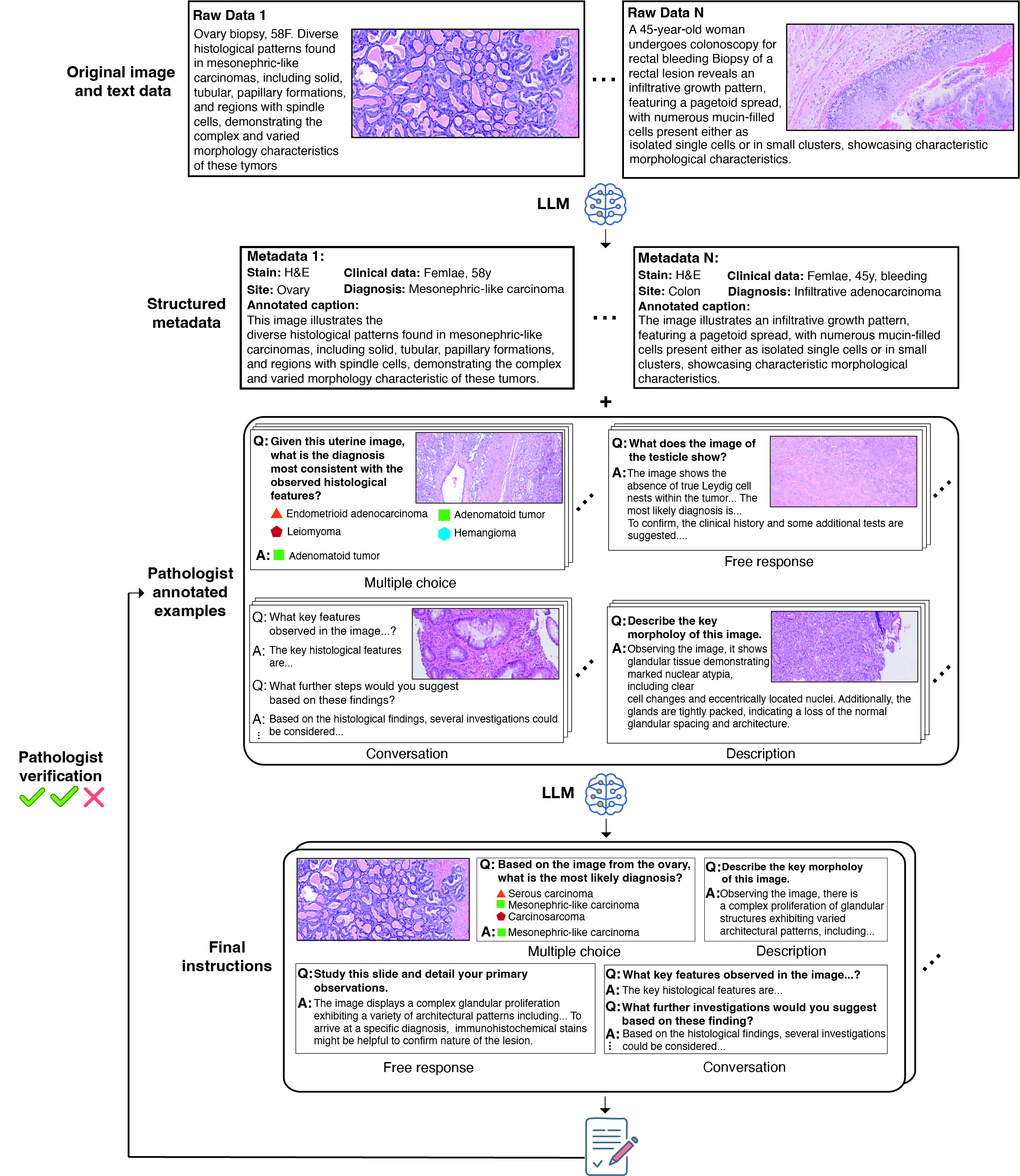}
    \captionof{figure}{\lucaedit{\textbf{LLM-assisted data-curation workflow for \pathchatnew{}.} 
    Raw pathology images and accompanying text reports are first parsed by a pretrained LLM to produce structured metadata (e.g., stain, clinical data, site, diagnosis, and an annotated caption). The LLM then drafts task-specific instructions using the \emph{image + metadata} together with a pool of pathologist-annotated exemplars; formats include multiple-choice, free-response, descriptive prompts, and short conversations. A board-certified pathologist reviews and edits all outputs; approved instructions are included in the manually verified examples for large-scale curation of task-specific, clinically grounded instructions, with periodic auditing by the pathologist.}}
    \label{fig:datacuration_workflow}
\end{center}

\captionsetup[table]{name=Extended Data Table,
labelfont=bf, 
}
\clearpage

\begin{table}[h]
\centering
\begin{tabular}{lccc}\\ \toprule     Model  & All-val & All-test-tiny & All-test\\ 	\midrule
Claude Sonnet 4& 0.548 (0.514, 0.584) & 0.610 (0.581, 0.637) & 0.577 (0.567, 0.587) \\
GPT-5 & 0.648 (0.612, 0.684) & 0.719 (0.692, 0.743) & 0.679 (0.670, 0.688) \\
GPT-5-mini & 0.617 (0.579, 0.652) & 0.683 (0.656, 0.711) & 0.648 (0.638, 0.657)\\
Gemini 2.5 pro & 0.668 (0.634, 0.702) & 0.714 (0.690, 0.738) & 0.679 (0.669, 0.688) \\
HuatuoGPT-Vision & 0.575 (0.536, 0.609) & 0.587 (0.561, 0.617) & 0.564 (0.554, 0.573) \\
LLaVA-OneVision & 0.428 (0.391, 0.465) & 0.489 (0.459, 0.518) & 0.452 (0.442, 0.463) \\
Llama-3.2 & 0.489 (0.449, 0.528) & 0.530 (0.501, 0.558) & 0.491 (0.481, 0.501) \\
PA-LLaVA & 0.376 (0.340, 0.415) & 0.422 (0.396, 0.451) & 0.389 (0.380, 0.399) \\
PathChat 1 & 0.579 (0.541, 0.617) & 0.640 (0.613, 0.668) & 0.609 (0.599, 0.619) \\
PathChat+ & 0.720 (0.685, 0.754) & 0.771 (0.747, 0.796) & 0.751 (0.742, 0.760) \\
Quilt-LLaVA & 0.416 (0.381, 0.454) & 0.441 (0.413, 0.470) & 0.406 (0.397, 0.416) \\
Qwen3-VL & 0.502 (0.466, 0.539)& 0.571 (0.543, 0.598)& 0.533 (0.522, 0.543)\\
Expert performance$\ast$ & - & 0.718 & - \\
\bottomrule
\end{tabular}
\caption{\kuanedit{\textbf{Performance on PathMMU multiple-choice questions, validation, test-tiny and test subset.} Accuracy is reported on subset of all validation (All-val, $n = 699$), all test tiny (All-test-tiny, $n = 1,150$) and all test (All-test, $n = 9,618$). 95\% confidence intervals from bootstrapping are included in parentheses. For more details see \textbf{MLLM evaluation} in \textbf{Methods}. $\ast$: The expert performance is quoted from the PathMMU paper\cite{sun2024pathmmu}. Note that the quoted expert performance is not directly comparable to all the evaluated model performance numbers since the Atlas subset has been updated by the PathMMU authors and is not the same as reported in the original paper.}} 
\label{tab:PathMMU_all}
\end{table}

\begin{table}[h]
    \centering
    \begin{tabular}{lcc}
    \toprule
    Model & Atlas (test-tiny) & Atlas (test-all) \\
    \midrule
    Claude Sonnet 4 & 0.591 (0.529, 0.659) & 0.630 (0.600, 0.660) \\
    GPT-5 & 0.769 (0.707, 0.827) & 0.740 (0.711, 0.766) \\
    GPT-5-mini & 0.716 (0.654, 0.779) & 0.712 (0.683, 0.741)\\
    Gemini 2.5 pro & 0.798 (0.740, 0.856) & 0.752 (0.724, 0.779) \\
    HuatuoGPT-Vision & 0.678 (0.611, 0.745) & 0.637 (0.607, 0.667) \\
    LLaVA-OneVision & 0.601 (0.538, 0.668) & 0.541 (0.511, 0.573) \\
    Llama-3.2 & 0.615 (0.553, 0.683) & 0.608 (0.578, 0.637) \\
    PA-LLaVA & 0.476 (0.409, 0.548) & 0.441 (0.409, 0.473) \\
    PathChat 1 & 0.788 (0.731, 0.841) & 0.753 (0.727, 0.779) \\
    PathChat+ & 0.827 (0.774, 0.875) & 0.840 (0.816, 0.862) \\
    Quilt-LLaVA & 0.495 (0.428, 0.567) & 0.476 (0.443, 0.506) \\
    Qwen3-VL & 0.534 (0.462, 0.606)  & 0.571 (0.540, 0.602)\\
    Expert performance$\ast$ & 0.683 & - \\
    \bottomrule
    \end{tabular}
    \caption{\kuanedit{\textbf{Performance on PathMMU multiple-choice questions, Atlas subset.} Accuracy is reported on test tiny subset ($n = 208$) and test subset ($n = 1,007$). 95\% confidence intervals from bootstrapping are included in parentheses. For more details see \textbf{MLLM evaluation} in \textbf{Methods}. $\ast$: The expert performance is quoted from the PathMMU study\cite{sun2024pathmmu}. Note that the quoted expert performance is not directly comparable to all the evaluated model performance numbers since the Atlas subset has been updated by the PathMMU authors and is not the same as reported in the original paper.}}
    \label{tab:PathMMU_Atlas}
\end{table}

\begin{table}[h]
    \centering
    \begin{tabular}{lcc}
    \toprule
    Model & EduContent (test-tiny) & EduContent (test-all) \\
    \midrule
    Claude Sonnet 4 & 0.667 (0.612, 0.718) & 0.633 (0.611, 0.654) \\
    GPT-5 & 0.757 (0.706, 0.812) & 0.689 (0.669, 0.709) \\
    GPT-5-mini & 0.757 (0.706, 0.812)& 0.669 (0.646, 0.691) \\
    Gemini 2.5 pro & 0.718 (0.659, 0.773) & 0.693 (0.672, 0.712) \\
    HuatuoGPT-Vision & 0.620 (0.557, 0.682) & 0.570 (0.548, 0.591) \\
    LLaVA-OneVision & 0.573 (0.514, 0.631) & 0.512 (0.489, 0.535) \\
    Llama-3.2 & 0.565 (0.506, 0.627) & 0.507 (0.488, 0.530) \\
    PA-LLaVA & 0.424 (0.361, 0.490) & 0.421 (0.399, 0.444) \\
    PathChat 1 & 0.678 (0.616, 0.737) & 0.631 (0.609, 0.651) \\
    PathChat+ & 0.769 (0.718, 0.820) & 0.735 (0.716, 0.755) \\
    Quilt-LLaVA & 0.498 (0.431, 0.557) & 0.437 (0.413, 0.460) \\
    Qwen3-VL & 0.612 (0.553, 0.671)  & 0.561 (0.539, 0.583) \\
    Expert performance$\ast$ & 0.690 & - \\ 
    \bottomrule
    \end{tabular}
\caption{\kuanedit{\textbf{Performance on PathMMU multiple-choice questions, EduContent subset.} Accuracy is reported on test tiny subset ($n = 255$) and test subset ($n = 1,938$). 95\% confidence intervals from bootstrapping are included in parentheses. For more details see \textbf{MLLM evaluation} in \textbf{Methods}. $\ast$: The expert performance is quoted from the PathMMU paper\cite{sun2024pathmmu}.}}
\label{tab:PathMMU_EduContent}
\end{table}

\begin{table}[h]
\centering
\begin{tabular}{lcc}\\ \toprule     Model  & PathCLS (test-tiny) & PathCLS (test-all)\\ 	\midrule
Claude Sonnet 4 & 0.350 (0.288, 0.418) & 0.333 (0.312, 0.356) \\
GPT-5 & 0.537 (0.463, 0.605) & 0.561 (0.537, 0.584) \\
GPT-5-mini & 0.503 (0.429, 0.576) & 0.513 (0.490, 0.537) \\
Gemini 2.5 pro & 0.525 (0.458, 0.599) & 0.561 (0.539, 0.585) \\
HuatuoGPT-Vision & 0.367 (0.299, 0.441) & 0.363 (0.341, 0.384) \\
LLaVA-OneVision & 0.226 (0.164, 0.288) & 0.239 (0.220, 0.261) \\
Llama-3.2 & 0.311 (0.249, 0.379) & 0.307 (0.287, 0.328) \\
PA-LLaVA & 0.305 (0.237, 0.379) & 0.286 (0.266, 0.307) \\
PathChat 1 & 0.424 (0.356, 0.497) & 0.452 (0.430, 0.475) \\
PathChat+ & 0.740 (0.678, 0.802) & 0.748 (0.728, 0.768) \\
Quilt-LLaVA & 0.299 (0.232, 0.367) & 0.271 (0.252, 0.293) \\
Qwen3-VL & 0.384 (0.316, 0.458)  & 0.376 (0.354, 0.398) \\
Expert performance & 0.789 & - \\
\bottomrule
\end{tabular}
\caption{\kuanedit{\textbf{Performance on PathMMU multiple-choice questions, PathCLS subset.} Accuracy is reported on test tiny subset ($n = 177$) and test subset ($n = 1,809$). 95\% confidence intervals from bootstrapping are included in parentheses. For more details see \textbf{MLLM evaluation} in \textbf{Methods}. $\ast$: The expert performance is quoted from the PathMMU study\cite{sun2024pathmmu}.}}
\label{tab:PathMMU_PathCLS}
\end{table}

\begin{table}[h]
    \centering
    \begin{tabular}{lcc}
    \toprule
    Model & SocialPath (test-tiny) & SocialPath (test-all) \\
    \midrule
    Claude Sonnet 4 & 0.668 (0.607, 0.729) & 0.625 (0.600, 0.646) \\
    GPT-5 & 0.738 (0.677, 0.790) & 0.694 (0.673, 0.717) \\
    GPT-5-mini & 0.668 (0.611, 0.729) & 0.655 (0.634, 0.677) \\ 
    Gemini 2.5 pro & 0.703 (0.646, 0.764) & 0.663 (0.641, 0.685) \\
    HuatuoGPT-Vision & 0.576 (0.511, 0.638) & 0.596 (0.575, 0.619) \\
    LLaVA-OneVision & 0.489 (0.419, 0.555) & 0.489 (0.464, 0.513) \\
    Llama-3.2 & 0.520 (0.454, 0.585) & 0.521 (0.499, 0.545) \\
    PA-LLaVA & 0.454 (0.393, 0.520) & 0.410 (0.385, 0.433) \\
    PathChat 1 & 0.638 (0.576, 0.699) & 0.614 (0.591, 0.637) \\
    PathChat+ & 0.751 (0.694, 0.808) & 0.717 (0.695, 0.738) \\
    Quilt-LLaVA & 0.441 (0.376, 0.507) & 0.454 (0.433, 0.478) \\
    Qwen3-VL & 0.642 (0.581, 0.703) & 0.561 (0.538, 0.582) \\
    Expert performance$\ast$ & 0.715 & - \\
    \bottomrule
    \end{tabular}
    \caption{\kuanedit{\textbf{Performance on PathMMU multiple-choice questions, SocialPath subset.} Accuracy is reported on test tiny subset ($n = 229$) and test subset ($n = 1,796$). 95\% confidence intervals from bootstrapping are included in parentheses. For more details see \textbf{MLLM evaluation} in \textbf{Methods}. $\ast$: The human expert performance is quoted from the PathMMU study\cite{sun2024pathmmu}.}}
    \label{tab:PathMMU_SocialPath}
\end{table}

\begin{table}[h]
    \centering
    \begin{tabular}{lcc}
    \toprule
    Model & PubMed (test-tiny) & PubMed (test-all) \\
    \midrule
    Claude Sonnet 4 & 0.687 (0.637, 0.744) & 0.641 (0.623, 0.657) \\
    GPT-5 & 0.747 (0.698, 0.797) & 0.713 (0.696, 0.729) \\
    GPT-5-mini & 0.719 (0.665, 0.769) & 0.688 (0.671, 0.704)\\
    Gemini 2.5 pro & 0.776 (0.730, 0.826) & 0.725 (0.709, 0.741) \\
    HuatuoGPT-Vision & 0.637 (0.584, 0.690) & 0.635 (0.619, 0.653) \\
    LLaVA-OneVision & 0.495 (0.438, 0.552) & 0.489 (0.471, 0.507) \\
    Llama-3.2 & 0.580 (0.520, 0.637) & 0.535 (0.517, 0.553) \\
    PA-LLaVA & 0.427 (0.367, 0.484) & 0.401 (0.384, 0.419) \\
    PathChat 1 & 0.633 (0.580, 0.687) & 0.636 (0.620, 0.654) \\
    PathChat+ & 0.769 (0.719, 0.815) & 0.753 (0.738, 0.768) \\
    Quilt-LLaVA & 0.438 (0.377, 0.498) & 0.416 (0.399, 0.433) \\
    Qwen3-VL &0.623 (0.566, 0.680) & 0.579 (0.561, 0.596) \\
    Expert performance$\ast$ & 0.729& - \\
    \bottomrule
    \end{tabular}
    \caption{\kuanedit{\textbf{Performance on PathMMU multiple-choice questions, Pubmed subset.} Accuracy is reported on test tiny subset ($n = 281$) and test subset ($n = 3,068$). 95\% confidence intervals from bootstrapping are included in parentheses. For more details see \textbf{MLLM evaluation} in \textbf{Methods}. $\ast$: The expert performance is quoted from the PathMMU paper\cite{sun2024pathmmu}.}}
    \label{tab:PathMMU_PubMed}
\end{table}
\begin{table}[h]
\centering
\begin{tabular}{lc}\\ \toprule     Model  & PathQABench MCQ\\ 	\midrule
Claude Sonnet 4 & 0.771 (0.686, 0.848) \\
GPT-5 & 0.848 (0.771, 0.914) \\
GPT-5-mini & 0.848 (0.711, 0.914)\\
Gemini 2.5 pro & 0.876 (0.810, 0.933) \\
HuatuoGPT-Vision & 0.571 (0.476, 0.667) \\
LLaVA-OneVision & 0.257 (0.171, 0.343) \\
Llama-3.2 & 0.733 (0.647, 0.810) \\
PA-LLaVA & 0.324 (0.238, 0.410) \\
PathChat 1 & 0.895 (0.829, 0.952) \\
PathChat+ & 0.943 (0.895, 0.981) \\
Quilt-LLaVA & 0.333 (0.248, 0.429) \\
Qwen3-VL & 0.638 (0.543, 0.724)\\
\bottomrule
\end{tabular}
\caption{\kuanedit{\textbf{Performance on multiple-choice questions of PathQABench MCQ.} Accuracy is reported on PathQABench MCQ ($n=105$) with 95\% confidence intervals from bootstrapping are included in parentheses. For more details see \textbf{MLLM evaluation} in \textbf{Methods}.}}
\label{tab:becnchmark_PathQABench}
\end{table}

\begin{table}[h]
    \centering
    \begin{tabular}{lccc}
    \toprule
    Model & BRACS & UniToPatho & HiCervix \\
    \midrule
    Claude Sonnet 4 & 0.230 (0.198, 0.263) & 0.252 (0.235, 0.270) & 0.249 (0.240, 0.259) \\
    GPT-5 & 0.349 (0.309, 0.388) & 0.263 (0.245, 0.281) & 0.456 (0.445, 0.466) \\
    GPT-5-mini & 0.340 (0.298, 0.381) & 0.392 (0.372, 0.411) & 0.426 (0.415, 0.436)\\
    Gemini 2.5 pro & 0.372 (0.333, 0.412) & 0.401 (0.381, 0.421) & 0.443 (0.433, 0.454) \\
    HuatuoGPT-Vision & 0.198 (0.167, 0.228) & 0.311 (0.294, 0.329) & 0.371 (0.360, 0.381) \\
    LLaVA-OneVision & 0.147 (0.118, 0.175) & 0.360 (0.341, 0.378) & 0.376 (0.366, 0.386) \\
    Llama-3.2 & 0.200 (0.170, 0.235) & 0.212 (0.196, 0.229) & 0.293 (0.282, 0.302) \\
    PA-LLaVA & 0.161 (0.132, 0.191) & 0.242 (0.226, 0.260) & 0.330 (0.320, 0.340) \\
    PathChat 1 & 0.558 (0.516, 0.596) & 0.506 (0.485, 0.526) & 0.368 (0.356, 0.379) \\
    PathChat+ & 0.633 (0.589, 0.672) & 0.522 (0.504, 0.541) & 0.715 (0.706, 0.724) \\
    Quilt-LLaVA & 0.160 (0.132, 0.189) & 0.207 (0.191, 0.222) & 0.372 (0.362, 0.382) \\
    Qwen3-VL & 0.232 (0.198, 0.265)& 0.250 (0.233, 0.266) & 0.258 (0.248, 0.267)\\
    \bottomrule
    \end{tabular}
    \caption{\kuanedit{\textbf{Performance on multiple-choice questions of BRACS, UniToPatho and HiCervix.} Accuracy is reported on BRACS ($n = 570$), UniToPatho ($n = 2,399$) and HiCervix ($n=8,051$). 95\% confidence intervals from bootstrapping are included in parentheses. For more details see \textbf{MLLM evaluation} in \textbf{Methods}.}
    \label{tab:benchmark_classification}}
\end{table}

\begin{table}[h]
\centering
\begin{tabular}{lc}\\ \toprule     Model  & PathQABench Caption\\ 	\midrule
Claude Sonnet 4 & 0.224 (0.212, 0.236) \\
GPT-5 & 0.242 (0.230, 0.253) \\
GPT-5-mini & 0.236 (0.224, 0.247)\\
Gemini 2.5 pro & 0.248 (0.235, 0.260) \\
HuatuoGPT-Vision & 0.217 (0.208, 0.226) \\
LLaVA-Med 1.5 & 0.200 (0.190, 0.209) \\
LLaVA-OneVision & 0.109 (0.094, 0.122) \\
Llama-3.2 & 0.192 (0.183, 0.201) \\
PathChat 1 & 0.263 (0.247, 0.280) \\
PathChat+ & 0.281 (0.270, 0.293) \\
Quilt-LLaVA & 0.199 (0.189, 0.209) \\
Qwen3-VL & 0.240 (0.227, 0.253)\\
\bottomrule
\end{tabular}
\caption{\kuanedit{\textbf{Performance on PathQABench Caption for image captioning.} METEOR score\cite{banerjee2005meteor} is reported on PathQABench Caption ($n=105$) with 95\% confidence intervals from bootstrapping are included in parentheses. For more details see \textbf{MLLM evaluation} in \textbf{Methods}.}
\label{tab:becnchmark_PathQABenchCaption}}
\end{table}

\begin{table}[ht]
\centering
\begin{tabular}{>{\raggedright}p{4.5cm} >{\raggedright\arraybackslash}p{12cm}}
\toprule
Site & Diagnoses \\
\midrule
Brain & Anaplastic Astrocytoma*, Ependymoma*, Glioblastoma*, Medulloblastoma*, Meningioma*, Oligodendroglioma*, Pilocytic Astrocytoma*, Pituitary Adenoma* \\
\midrule
Breast & Invasive Breast Carcinoma of No Special Type (NST), Invasive Lobular Carcinoma,  Mixed Ductal and Lobular Carcinoma \\
\midrule
Endocrine & Adrenocortical Carcinoma*, Anaplastic Thyroid Carcinoma*, Follicular Thyroid Carcinoma*, Medullary Thyroid Carcinoma*, Papillary Thyroid Carcinoma \\
\midrule
GI & Colorectal Adenocarcinoma, Esophageal Adenocarcinoma*, Esophageal Squamous Cell Carcinoma*, Gastric Adenocarcinoma, Gastric Signet Ring Cell Carcinoma*, Squamous Cell Carcinoma of the Anus* \\ 
\midrule
GYN & Clear Cell Carcinoma of the Ovary*, Clear Cell Carcinoma of the Uterus*, Endometrioid Carcinoma of the Ovary*, Endometrioid Carcinoma of the Uterus, High-Grade Serous Carcinoma of the Ovary*, Low-Grade Serous Carcinoma of the Ovary*, Mucinous Carcinoma of the Ovary*, Serous Borderline Tumor of the Ovary*, Serous Carcinoma of the Uterus*, Squamous Cell Carcinoma of the Cervix* \\
\midrule
Lung & Adenocarcinoma of the Lung, Atypical Carcinoid Tumor*, Large Cell Neuroendocrine Carcinoma*, Mixed Large Cell and Small Cell Neuroendocrine Carcinoma, Small Cell Carcinoma of the Lung*, Squamous Cell Carcinoma of the Lung, Typical Carcinoid Tumor* \\
\midrule
Male Reproductive Tract & Prostatic Adenocarcinoma, Seminoma* \\ 
\midrule
Pancreaticohepatobiliary & Hepatocellular Carcinoma*, Pancreatic Adenocarcinoma*, Pancreatic Neuroendocrine Tumor (PanNET)* \\ 
\midrule
Skin and Connective Tissue & Cutaneous Squamous Cell Carcinoma, Malignant Melanoma, Merkel Cell Carcinoma* \\
\midrule
Urinary Tract & Adenocarcinoma of the Bladder*, Bladder Urothelial Carcinoma, Chromophobe Renal Cell Carcinoma*, Clear Cell Renal Cell Carcinoma*, Nephroblastoma (Wilms Tumor)*, Papillary Renal Cell Carcinoma*, Renal Oncocytoma*, Upper Tract Urothelial Carcinoma* \\
\bottomrule
\end{tabular}
\caption{\textbf{Unique diagnoses by tissue site in DDxBench.} GI: Gastrointestinal, GYN: Gynecology. \lucaedit{Rare diseases, as defined by an incidence of fewer than 6 in 100,000 per year, are marked with an asterisk (*)\cite{aslaniIncidenceTrendsGastric2024, chenIncidenceDemographicsSurvival2021, fengFrequencyIncidenceSurvival2019, gudbjartssonRenalOncocytomaClinicopathological2005, ingimarssonChromophobeRenalCell2011, limTrendsThyroidCancer2017, liTrendsIncidenceRates2003, patel2018incidence, runarssonEpidemiologicalClinicopathologicalStudy2024, upretyTrendsIncidenceSurvival2025, wijayabahuUterineCancerIncidence2024}.}}
\label{tab:diagnoses_by_site}
\end{table}

\begin{table}[h]
\centering
\setlength{\tabcolsep}{6pt}
\begin{tabular}{l c l c c}
\toprule
\textbf{Agent LLMs} & \textbf{Multi-Agent} & \textbf{Captioner} &
\textbf{Primary (top-1)} & \textbf{Primary \& Differential (top-3)} \\
\midrule
GPT-5-mini & $\checkmark$ & PathChat+ &
0.860 (0.800, 0.907) & 0.927 (0.880, 0.967) \\
GPT-4.1 & $\checkmark$ & PathChat+ &
0.813 (0.753, 0.873) & 0.927 (0.880, 0.967) \\
GPT-5-mini & $\times$ & PathChat+ &
0.780 (0.707, 0.847) & 0.880 (0.827, 0.927) \\
GPT-5-mini & $\checkmark$ & PathChat 1 &
0.640 (0.560, 0.713) & 0.807 (0.740, 0.867) \\
GPT-5-mini & $\checkmark$ & GPT-5-mini &
0.427 (0.353, 0.507) & 0.627 (0.547, 0.700) \\
\bottomrule
\end{tabular}
\caption{\lucaedit{\textbf{Ablations of \agent{} on DDxBench.} Each row ablates one component of the full system to isolate its effect. “Agent LLMs” is the backbone used by the supervisor and explorer agents for planning/reasoning; “Captioner” is the model used to extract morphology and to produce the final differential (\pathchatnew{} unless ablated). The first row is the reference system for \agent{}: a supervisor–explorer \emph{multi-agent} setup with GPT-5-mini agents and \pathchatnew{} as the captioner. Row 2 swaps the agent backbone to GPT-4.1 while keeping the hierarchy and captioner fixed (tests value of a reasoning agent). Row 3 collapses the hierarchy to a single explorer (no supervisor; Multi-Agent $\times$) with the same agent backbone and captioner (tests the supervisor–explorer design). Rows 4–5 keep the agent side fixed and ablate the captioner: replacing \pathchatnew{} with PathChat 1 (row 4) or a general-purpose GPT-5-mini captioner (row 5). Values are accuracy (higher is better) for primary (top-1) and primary + differentials (top-3) with 95\% CIs; $\checkmark$ indicates the supervisor–explorer hierarchy is present, $\times$ indicates the single-agent ablation.}}\label{tab:agent-perf}
\end{table}

\begin{table}[]
\centering
\begin{tabular}{lcc}\\ \toprule     Model  &  Primary Diagnosis & Primary + Add. Differentials\\ 	\midrule
Claude Sonnet 4 & 0.427 (0.347, 0.507) & 0.560 (0.487, 0.640)\\ 
GPT-5 & 0.407 (0.327, 0.487) & 0.613 (0.533, 0.700)\\
GPT-5-mini & 0.433 (0.353, 0.513) & 0.673 (0.600, 0.745)\\
Gemini 2.5 pro & 0.513 (0.433, 0.593) & 0.713 (0.647, 0.787)\\
HuatuoGPT-Vision & 0.333 (0.260, 0.413) & 0.540 (0.460,0.620) \\
LLaVA-OneVision & 0.107 (0.067, 0.160) & 0.167 (0.113,0.227) \\
PathChat 1 & 0.720 (0.647, 0.787) &  0.887 (0.833,0.933)\\
\pathchatnew{} & 0.800 (0.740, 0.860) & 0.920 (0.873,0.960) \\
Qwen3-VL &0.387 (0.313, 0.467) & 0.627 (0.553,0.707)) \\
\bottomrule
\end{tabular}
\caption{\kuanedit{\textbf{Performance of \pathchatnew{} with expert-curated ROIs on DDxBench.} Primary Diagnosis (top-1 accuracy) and Primary + Differential (top-3 accuracy) are reported on DDxBench ($n=150$) Each slide is represented by ten curated ROIs, which are directly fed to \pathchatnew{}. 95\% confidence intervals from bootstrapping are included in parentheses. For more details see \textbf{MLLM evaluation} in \textbf{Methods}.}}
\label{tab:ddxbench-roi}
\end{table}

\begin{table}[h]
  \centering
  \begin{tabular}{p{7.5cm}|p{3cm}}
    \toprule
    Hyperparameter & Value \\
    \midrule
    Automatic mixed precision & BF16 \\
    DeepSpeed ZeRO & Stage 3 \\
    Batch size & 128 \\
    Learning rate scheduler & Cosine \\
    Warmup ratio & 0.03 \\
    Peak learning rate & 2e-3 \\
    AdamW $\beta$ & (0.9, 0.999) \\
    AdamW $\epsilon$ & 1e-8 \\
    Weight decay & 0. \\
    Gradient clipping max. norm & 1.0 \\
    Training epochs & 1 \\
    Gradient checkpointing & Yes \\
    TF32 & Yes \\
    \bottomrule
  \end{tabular}
  \caption{\textbf{Hyperparameters for \pathchatnew{} pretraining.} Training was conducted on 8 $\times$ 80GB NVIDIA A100 GPUs. The reported batch size is the effective batch size across all GPUs. The learning rate was linearly increased from zero to the peak value over $\textit{warmup ratio} \times \textit{total batches}$ steps, then decayed to zero following a cosine schedule. These parameters closely follow the original pretraining of PathChat 1.}
  \label{tab:hparams_pretrain}
\end{table}

\begin{table}[h]
  \centering
  \begin{tabular}{p{7.5cm}|p{3cm}}
    \toprule
    Hyperparameter & Value \\
    \midrule
    Automatic mixed precision & BF16 \\
    DeepSpeed ZeRO & Stage 3 \\
    Batch size per GPU & 2 \\
    Gradient accumulation steps & 2 \\
    Learning rate scheduler & Cosine \\
    Warmup ratio & 0.03 \\
    Peak learning rate & 2e-5 \\
    AdamW $\beta$ & (0.9, 0.999) \\
    AdamW $\epsilon$ & 1e-8 \\
    Weight decay & 0. \\
    Gradient clipping max. norm & 1.0 \\
    Training epochs & 1 \\
    Gradient checkpointing & Yes \\
    TF32 & Yes \\
    \bottomrule
  \end{tabular}
  \caption{\textbf{Hyperparameters for \pathchatnew{} multimodal LLM finetuning.} Training was performed on 24 $\times$ 80GB NVIDIA A100 GPUs across 3 compute nodes using multi-node distributed training. The reported batch size is per GPU; the effective batch size is given by (\textit{batch size} $\times$ \textit{number of GPUs} $\times$ \textit{gradient accumulation steps}). The learning rate was linearly increased from zero to the peak value over $\textit{warmup ratio} \times \textit{total batches}$ steps, then decayed to zero following a cosine schedule.}
  \label{tab:hparams_finetune}
\end{table}



\begin{nolinenumbers}
\clearpage
\section*{References} 
\vspace{2mm}

\begin{spacing}{0.9}
\bibliographystyle{naturemag}
\bibliography{sample}
\end{spacing}
\end{nolinenumbers}

\end{document}